%% file: main.tex
\pdfoutput=1

\documentclass[11pt]{article}
\PassOptionsToPackage{usenames,dvipsnames}{xcolor}

\usepackage[]{acl}

\usepackage{times}
\usepackage{latexsym}

\usepackage[T1]{fontenc}

\usepackage[utf8]{inputenc}

\usepackage{microtype}

\usepackage{amsmath}
\usepackage{amssymb}
\usepackage{booktabs}
\usepackage{subcaption}

\newcommand{\norm}[1]{\left\lVert#1\right\rVert}
\usepackage{algorithm} 
\usepackage[noend]{algorithmic} 

\usepackage{todonotes}
\usepackage[usenames,dvipsnames]{xcolor}
\usepackage{bm}  
\usepackage{enumitem}
\usepackage{graphicx}

\input{math_commands.tex}

\title{Finding Structural Knowledge in Multimodal-BERT}

\author{Victor Milewski$^1$ \and Miryam de Lhoneux$^{1,2,3}$ \and Marie-Francine Moens$^1$\\
  $^1$Department of Computer Science, KU Leuven\\
  $^2$Department of Computer Science, University of Copenhagen\\
  $^3$Department of Linguistics and Philology, Uppsala University\\
  \texttt{victor (dot) milewski (at) kuleuven.be}}

\date{\today}

\begin{document}
\maketitle
\begin{abstract}
In this work, we investigate the knowledge learned in the embeddings of multimodal-BERT models.
More specifically, we probe their capabilities of storing the grammatical structure of linguistic data and the structure learned over objects in visual data. 
To reach that goal, we first make the inherent structure of language and visuals explicit by a dependency parse of the sentences that describe the image and by the dependencies between 
the object regions in the image, respectively. 

We call this explicit visual structure the \textit{scene tree}, 
that is based on the dependency tree of the language description. 
Extensive probing experiments show that the multimodal-BERT models 
do not encode these scene trees.
Code available at \url{https://github.com/VSJMilewski/multimodal-probes}.
\end{abstract}
\setlength{\textfloatsep}{5pt}
\setlength{\floatsep}{0pt}
\section{Introduction}
In recent years, contextualized embeddings have become increasingly important. 
Embeddings created by the BERT model and its variants have been used to get state-of-the-art performance in many tasks \citep{devlin-etal-2019-bert, liu2019roberta, yang2019xlnet, radford2018gpt, radford2019gpt2,brown2020gpt3}. 
Several multimodal-BERT models have been developed that learn multimodal contextual embeddings through training jointly on linguistic data and visual data \citep{lu2019vilbert, su2019vlbert, li2019visualbert, chen2020uniter}. 
They achieve state-of-the-art results across many tasks and benchmarks, such as Visual Question Answering \citep{balanced_vqa_v2}, image and text retrieval \citep{lin2014microsoft}, and Visual Commonsense Reasoning \citep{suhr2018corpus}.\footnote{From here on we refer to the text-only BERT models as 'BERT' and the multimodal-BERT models as 'multimodal-BERTs'.}
 
BERT and multimodal-BERTs are blackbox models that are not easily interpretable.  
It is not trivial to know what knowledge is encoded in the models and their embeddings.  
A common method for getting insight into the embeddings of both textual and visual content is probing.

\begin{figure}[t]
    \centering
    \includegraphics[width=\linewidth]{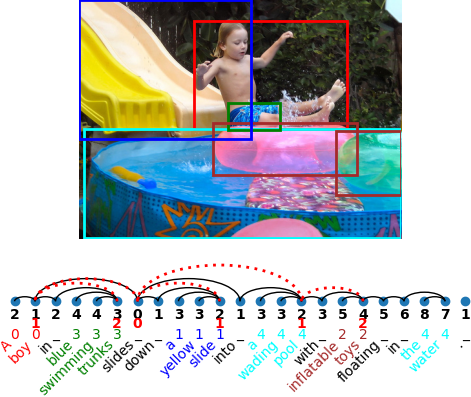}
    \caption{Example of the mapping from the linguistic dependency tree to the visual tree. The borders of the regions in the image have the same color as the phrase they are attached to. The rows below the image are the textual tree depth (in black), the visual tree depth (in red), the phrase index, and the words in the sentence.}
    \label{fig:tree_mapping}
\end{figure}

Language utterances have an inherent grammatical structure that contributes to their meaning.
Natural images have a characteristic spatial structure that likewise allows humans to interpret their meaning. 
In this paper we hypothesize that the textual and visual embeddings learned from images that are paired with their descriptions encode structural knowledge of both the language and the visual data.
Our goal is to reveal this structural knowledge with the use of probing.
More specifically, in order to perform this probing, we first make the inherent structure of language and visuals explicit by a mapping between a dependency parse of the sentences that describe the image and by the dependency between the object regions in the image, respectively.
Because the language truthfully describes the image, and inspired by \citet{draschkow2017scene}, we define a visual structure that correlates with the dependency tree structure and that arranges object regions in the image in a tree structure.
We call this visual dependency tree the \textit{scene tree}.  
An example of this mapping to the scene tree is visualized in Figure~\ref{fig:tree_mapping}.

The aligned dependency tree and scene tree allow us to 
conduct a large set of experiments aimed at discovering encoded structures in neural representations obtained from multimodal-BERTs. 
By making use of the structural probes proposed by \citet{hewitt-manning-2019-structural}, we compare the dependency trees learned by models with or without provided image features. 
Furthermore, we investigate if scene trees are learned in the object region embeddings. 

\paragraph{Research Questions}
In this study, we aim to answer the following research questions.
\begin{itemize}
    \item \textbf{RQ 1:} Do the textual embeddings trained with a multimodal-BERT retain their structural knowledge? \\
    \textbf{Sub-RQ 1.1:} To what extent does the joint training in a multimodal-BERT influence the structures learned in the textual embeddings?
    \item \textbf{RQ 2:} Do the visual embeddings trained with a multimodal-BERT learn to encode a scene tree?
\end{itemize}

In a broader framework this study might contribute to better representation learning inspired by how humans acquire language in a perceptual context.
It stimulates the learning of representations that are compositional in nature and are jointly influenced by the structure of language and the corresponding structure of objects in visuals.  

\section{Related Work}
\paragraph{Probing studies}
Several studies have been performed that aim at analyzing BERT and multimodal-BERTs. For BERT, probes are designed that explore gender bias \citep{bhardwaj2021probe_gender_bias}, 
relational knowledge \citep{singh-etal-2020-bertnesia}, 
linguistic knowledge for downstream tasks \citep{liu-etal-2019-linguistic}, 
part-of-speech knowledge \citep{hewitt2019designing,hewitt-etal-2021-conditional},
and for sentence and dependency structures \citep{tenney2019structure,hewitt-manning-2019-structural}. 
These studies have shown that BERT latently learns to encode linguistic structures in its textual embeddings. 
\citet{basaj2021explaining} made a first attempt at converting the probes to the visual modality and evaluated the information stored in the features created by visual models trained with self-supervision. 

For multimodal-BERTs, one study by \citet{parcalabescu2020seeing} investigates how well these models learn to count objects in images and how well they generalize to new quantities. 
They found that the multimodal-BERTs overfit the dataset bias and fail to generalize to out-of-distribution quantities. 
\citet{frank-etal-2021-vision} found that visual information is much more used for textual tasks than textual information is used for visual tasks when using multimodal models.
These findings suggest more needed research into other capabilities of and knowledge in multimodal-BERT embeddings. 
We build on this line of work but aim to discover structures encoded in the textual and visual embeddings learned with multimodal-BERTs. This is a first step towards finding an aligned structure between text and images. Future work could exploit this to make textual information more useful for visual tasks.

\paragraph{Structures in visual data}
There is large research interest in identifying structural properties of images e.g., scene graph annotation of the visual genome dataset \citep{krishnavisualgenome}. 
In the field of psychology, research towards scene grammars \citep{draschkow2017scene} evidences that humans assign certain grammatical structures to the visual world.
Furthermore, some studies investigate the grounding of textual structures in images, 
such as syntax learners \citep{shi_visually_2019} and visually grounded grammar inducers \cite{zhao_visually_2020}. Here the complete image is used, without considering object regions and their composing structure, to aid in predicting linguistic structures.

Closer to our work, \citet{Elliott2013ImageDU} introduced visual dependency relations (VDR), where spatial relations are created between object in the image. The VDR can also be created by locating the object and subject in a caption and matching it with object annotations in the image \citep{Elliott2015DescribingIU}. Our scene tree differs, since it makes use of the entire dependency tree of the caption to create the visual structure.

\section{Background}
\paragraph{Multimodal-BERT}
Many variations of the BERT model implement a transformer architecture to process both visual and linguistic data, e.g., images and sentences. These Multimodal-BERTs can be categorized into two groups: single-stream and dual-stream encoders. In the former, a regular BERT architecture processes the concatenated input of the textual description and the image through a transformer stack. 
This allows for an "unconstrained fusion of cross-modal features" \citep{bugliarello2021volta}. Some examples of these models are ViL-BERT \citep{su2019vlbert},  VisualBERT \citep{li2019visualbert}, and UNITER \citep{chen2020uniter}.

In the dual-stream models, the visual and linguistic features are first processed separately by different transformer stacks, followed by several transformer layers with alternating \textit{intra-modal} and \textit{inter-modal} interactions. 
For the \textit{inter-modal} interactions, the query-key-value matrices modeling the multi-head self-attention are computed, and then the key-value matrices are exchanged between the modalities. 
This limits the interactions between the modalities but increases the expressive power with separate parameters. Examples of such dual-stream models are ViLBERT \citep{lu2019vilbert}, LXMERT \citep{tan2019lxmert}, and ERNIE-ViL \citep{yu2021ernie}.\footnote{The ERNIE-ViL model is trained with scene graphs of the visual genome dataset. We do not probe this model as there is an overlap between the training data of ERNIE-ViL and our evaluation data.}

\section{Method}
\subsection{Tree Structures}\label{sec:tree}
In the probing experiments we assume that the structural knowledge of a sentence is made explicit by its dependency tree structure and that likewise the structural knowledge of an image is represented by a tree featuring the dependencies between object regions. 
Further, we assume that the nodes of a tree (words in the dependency tree of the sentence, phrase labels in the region dependency tree of the image) are represented as embeddings obtained from a layer in BERT or in a multimodal-BERT. 

To generate the depths and distances values from the tree, we use properties of the embedding representation space \citep{mikolov2013distributed}. 
For example, similar types of relations between embeddings have a similar distance between them, such as counties and their capital city. 
The properties we use are that the length (the norm) of a vector which describes the depth in a tree and the distance between nodes that can be translated as the distance between vectors. 

\paragraph{Generating distance values}
For the distance labels, a matrix $\mD \in \sN^{n\times n}$ is required, with each $\mD_{ij}$ describing the distance between nodes $i$ and $j$.
To fill the matrix, we iterate over all possible pairs of nodes. 
For nodes $i$ and $j$, it is computed by starting at node $i$ in the tree and traverse it until node $j$ is reached while ensuring a minimum distance. 
This is achieved by using the breadth-first search algorithm. 

\paragraph{Generating depth values}
For the depth labels, we generate a vector $\vd \in \sN^n$, with $n$ the number of nodes in the tree.
There is a single node that is the root of the tree, to which we assign a depth of zero. 
The depth increases at every level below.

\subsection{Constructing the Trees}\label{sec:data_processing}
\begin{algorithm}[pt]
\caption{$ConstructSceneTree(T_t, P, I)$}
\label{alg:tree_mapping}
\begin{algorithmic}[1] 
\REQUIRE Language dependency tree $T_t = \{E_t,V_t\}$, with $V_t$ the set of $TextIDs$ for words in a sentence and $E_t$ the set of edges such that each $e_{t} = (v_{t,j}, v_{t,k})$, where $v_{t,k}$ is a child node of $v_{t,j}$
\REQUIRE Set of phrases $P$, each $p_i$ describes one or more regions and covers multiple words
\REQUIRE Image $I$
\ENSURE Scene tree $T_s$
\STATE $V_s = \{\}$, set of Nodes in Scene Tree $T_s$
\STATE $E_s= \{\}$, set of Edges in Scene Tree $T_s$
\STATE $v_{s,0} = I$, set Image as root node
\STATE $D_0 = 0$, set root node depth as 0
\STATE $add(V_s, v_{s,0})$
\STATE $v_{t,0} = FindRootNode(T_t)$
\STATE $PhraseID2TextID(0) = v_{t,0}$
\FOR{$p_i \in P$}
    \STATE $v_{t,k} = FindHighestNode(p_i)$
    \STATE $PhraseID2TextID(p_i) = v_{t,k}$
    \STATE $D_i = DepthInTree(T_t, v_{t,k})$
\ENDFOR
\FOR{$p_i \in P$ \textbf{ordered by} $D$}
    \STATE $v_{t,k} = PhraseID2TextID(p_i)$
    \WHILE{\textbf{True}}
        \STATE $e_{t} = EdgeWithChildNode(E, v_{t,k})$
        \STATE $v_{t,j} = SelectParentNode(e_{t})$
        \STATE $p_p = TextID2PhraseID(v_{t,j})$
        \IF{$p_p \in V_s$}
            \STATE $add(V_s, p_i), \quad add(E_s, (p_p, p_i))$
            \STATE $D_i = D_p + 1$
            \STATE \textbf{break while loop}
        \ELSE
            \STATE $v_{t,k} = v_{t,j}$
        \ENDIF
    \ENDWHILE
\ENDFOR
\RETURN $T_s$
\end{algorithmic}
\end{algorithm}
\paragraph{Language dependency tree}
We use the dependency tree as linguistic structure. The tree annotations are according to the Stanford dependency guidelines \citep{de2008stanford}. 
They can either be provided as gold-standard in the dataset, or generated using the spacy dependency parser \citep{spacy}.

\paragraph{Scene tree}
\citet{draschkow2017scene} found that there are commonalities between words in language and objects in scenes, allowing to construct a scene grammar. 
Furthermore, \citet{zhao_visually_2020} have shown that an image provides clues that improve grammar induction. 
In line with these works, we want a visual structure that aligns with a linguistic representation like the dependency tree. 

As visual structure, a scene graph could be used for the relations between regions \citep{krishnavisualgenome}. 
However, the unconstrained graph 
is difficult to align with the dependency tree.
Therefore, we propose a novel visual structure, the \textit{scene tree}, that is created by mapping a textual dependency tree to the object regions of an image.
An example of such a mapping for an image-sentence pair is given in Figure~\ref{fig:tree_mapping}.  
This process requires a tree for the sentence and paired data for images and sentences. 

Each node in the scene tree directly matches one or more visual regions. 
The node description is a phrase that covers multiple words in the sentence (or nodes in the dependency tree).
The output of this method is a tree that contains the phrase trees that directly correspond to the regions.
The algorithm is completely described as pseudo-code in Algorithm~\ref{alg:tree_mapping}.

The algorithm starts by initializing the scene tree. 
We set the full image as the root node. 
For each phrase that describes an image region, we select the dependency tree node (or word with a $TextID$) that is closest to the root and assign this a phrase ID. This creates a mapping between the phrases (Phrase IDs) and dependency tree nodes (Text IDs) $PhraseID2TextID$, and its reverse $TextID2PhraseID$. 
We assign each phrase an initial depth, based on the word it maps to in $PhraseID2TextID$. 
On line 12, the loop over the phrases that describe the object regions starts, to find the direct parent for each phrase so it can be added to the new scene tree.
For each phrase $p_i$, we select the matching dependency tree node the $v_{t,k}$ from $PhraseID2TextID$. From $v_{t,k}$ we follow the chain of parent nodes, until an ancestor $v_{t,l}$ is found that points back to a phrase $p_j$ (using $TextID2PhraseID$) that is already a member of the scene tree. Phrase $p_i$ is added to the tree as child of $p_j$.
The completed tree of phrases is our \textit{scene tree}.

\subsection{Embeddings}\label{sec:embeddings}
\paragraph{Textual embeddings}
For each sentence $l$, every word becomes a node $n_i$ in the tree, such that we have a sequence of $s$ nodes $n_{1:s}^l$. To obtain the textual embeddings $\vh_{1:s}^l \in \sR^{m}$, we do a wordpiece tokenization \citep{wu2016google} and pass the sentence into BERT. Depending on the requested layer, we take the output of that BERT layer as the embeddings. For nodes with multiple embeddings because of the wordpiece tokenization, we take the average of those embeddings.

To obtain the textual embeddings $\vh_{1:s}^l$ for a multimodal-BERT, we use the same process but also provide visual features. 
When an image is present, we enter the visual features (as described in the next paragraph), otherwise, a single masked all-zero feature is entered.

\paragraph{Visual embeddings}
For sentence with image $l$, the sequence of $s$ nodes $n_{1:s}^l$ consists of the number of regions plus the full image. 
The visual embeddings $\vh_{1:s}^l \in \sR^{m}$ are obtained by passing the raw Faster R-CNN features \citep{ren2015faster} into the multimodal-BERT. 
Depending on the requested layer, we take the output of that multimodal-BERT layer as the embeddings.

\subsection{Structural Probes}\label{sec:probes}
Here we shortly describe the structural probes as defined by \citet{hewitt-manning-2019-structural}. 
Originally designed for text, we use these probes to map from an embedding space (either textual embeddings or visual embeddings) to depth or distance values as defined in Section~\ref{sec:tree}.

\paragraph{Distance probe}
Given a sequence of $s$ nodes $n_{1:s}^l$ (words or objects) and their embeddings $\vh_{1:s}^l \in \sR^{m}$, where $l$ identifies the sequence and $m$ the embedding size, we predict a matrix of $s\times s$ distances.
First, we define a linear transformation $\mB \in \mathbb{R}^{k\times m}$ with $k$ the probe rank, such that $\mB^T\mB$ is a positive semi-definite, symmetric matrix. 
By first transforming a vector $\vh$ with matrix $\mB$, we get its norm like this: $(\mB\vh)^T(\mB\vh)$.
To get the squared distance between two nodes $i$ and $j$ in sequence $l$, we compute the difference between node embeddings $\vh_i$ and $\vh_j$ and take the norm following equation~\ref{eq:distance}:
\begin{equation}
\mD_{ij} = (\mB(\vh_i^l-\vh_j^l))^T(\mB(\vh_i^l-\vh_j^l))\label{eq:distance}
\end{equation}
The only parameters of the distance probe are now the transformation matrix $\mB$, which can easily be implemented as a fully connected linear layer.
Identical to the work by \citet{hewitt-manning-2019-structural}, the probe is trained through stochastic gradient descent. 

\paragraph{Depth probe}
For the depth probe, we transform the embedding of each node $n_i$ to their norm, so we can construct the vector $\vd$.
This imposes a total order on the elements and results in the depths. 
We compute the squared vector norm $\norm{\vh_i}_\mB^2$ with the following equation:
\begin{equation}
\vd_i = \norm{\vh_i}_\mB^2 = (\mB\vh_i^l)^T(\mB\vh_i^l)\label{eq:depth}
\end{equation}

\section{Experimental Setup}
\subsection{Data}\label{sec:data}
By using a text-only dataset, we can test how the textual embeddings of the multimodal-BERTs perform compared to the BERT model, without the interference from the visual embeddings. 
This allows us to see how much information the multimodal-BERTs encode in the visual embeddings. 

Therefore, we use the Penn Treebank (PTB3) \citep{marcus1999treebank}. 
It is commonly used for dependency parsing (also by \citet{hewitt-manning-2019-structural} from whom we borrow the probes) and consists of gold-standard dependency tree annotations according to the Stanford dependency guidelines \citep{de2008stanford}. 
We use the default training/validation/testing split, that is, the subsets 2-21 for training, 22 for validation and 23 for testing 
of the Wall Street Journal sentences. 
This provides us with 39.8k/1.7k/2.4k sentences for the splits, respectively. 

The second dataset is the Flickr30k dataset \citep{young2014flickr30k}, which consists of multimodal image captioning data.
It has five caption annotations for each of the 30k images. An additional benefit of this dataset are the existing extensions, specifically the Flickr30k-Entities (F30E) \citep{plummer2015flickr30k_entities}. 
In F30E all the phrases in the captions are annotated and match with region annotations in the image. 
This paired dataset is used to create the scene trees proposed in Section~\ref{sec:data_processing}.

The Flickr30k dataset does not provide gold-standard dependency trees.
Therefore, the transformer based Spacy dependency parser \citep{spacy} is used to generate silver-standard dependency trees according to the Stanford dependency guidelines \citep{de2008stanford}.  
The dataset consists of 30k images, with (mostly) 5 captions each, resulting in 148.9k/5k/5k sentences for the training/validation/testing splits, respectively. 

\subsection{Models}\label{sec:models}
We use two different multimodal-BERTs, one \textbf{single-stream} and one \textbf{dual-stream} model. 
As implementation for the multimodal-BERTs, we make use of the \textsc{Volta} library \citep{bugliarello2021volta}. 
Here, all the models are implemented and trained under a controlled and unified setup with regard to hyperparameters and training data. 
Based on the performance under this unified setup on the Flickr30k image-sentence matching task, we have chosen the best performing models: ViLBERT \citep{lu2019vilbert} as single-stream model and UNITER \citep{chen2020uniter} as dual-stream model.

When probing the textual embeddings, we also use a text-only \textbf{BERT-base model} (from here on referred to as BERT) \citep{devlin-etal-2019-bert}.
\citet{hewitt-manning-2019-structural} use the same model, allowing for easy comparability. 
The implementation used is from the HuggingFace Transformer library \citep{wolf-etal-2020-transformers}.

\paragraph{Hyperparameters}
For our setup and metrics, we follow the setup from \citet{hewitt-manning-2019-structural}. 
The batch size is set to 32 and we train for a maximum of 40 epochs. 
Early stopping is used to terminate training after no improvement on the validation L1-loss for 5 epochs. 

\subsection{Metrics}\label{sec:metrics}
\begin{figure*}[t]
    \centering
    \begin{subfigure}[t]{0.9\linewidth}
        \centering
        \includegraphics[scale=0.4]{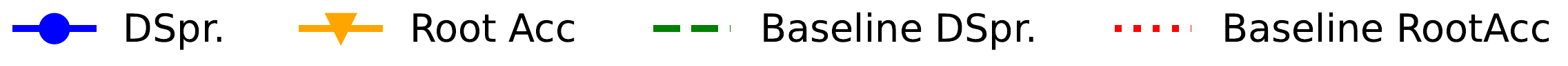}
    \end{subfigure}
    \\
    \begin{subfigure}[t]{0.32\linewidth}
        \centering
        \includegraphics[scale=0.48]{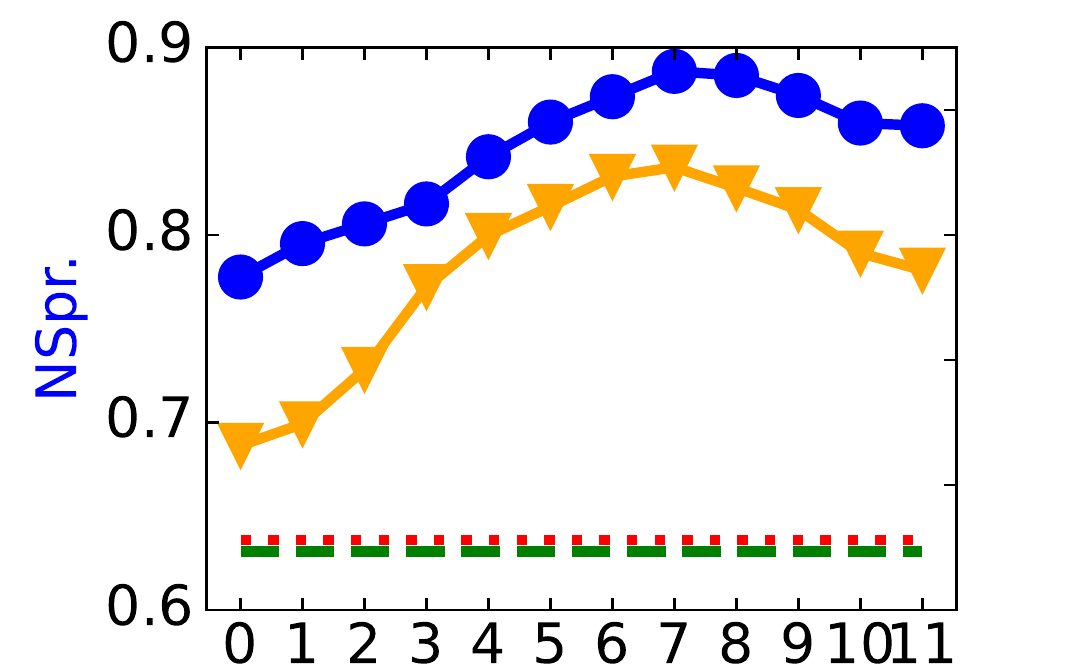}
        \caption{BERT}
        \label{fig:ptb_depth_bert}
    \end{subfigure}
    \begin{subfigure}[t]{0.32\linewidth}
        \centering
        \includegraphics[scale=0.48]{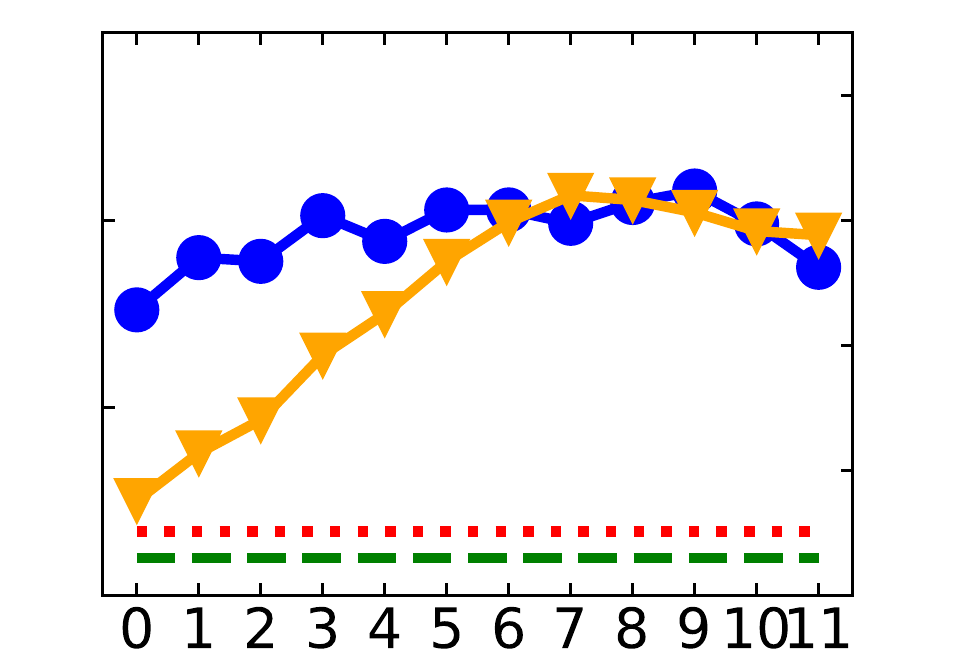}
        \caption{UNITER}
        \label{fig:ptb_depth_unit}
    \end{subfigure}
    \begin{subfigure}[t]{0.32\linewidth}
        \centering
        \includegraphics[scale=0.48]{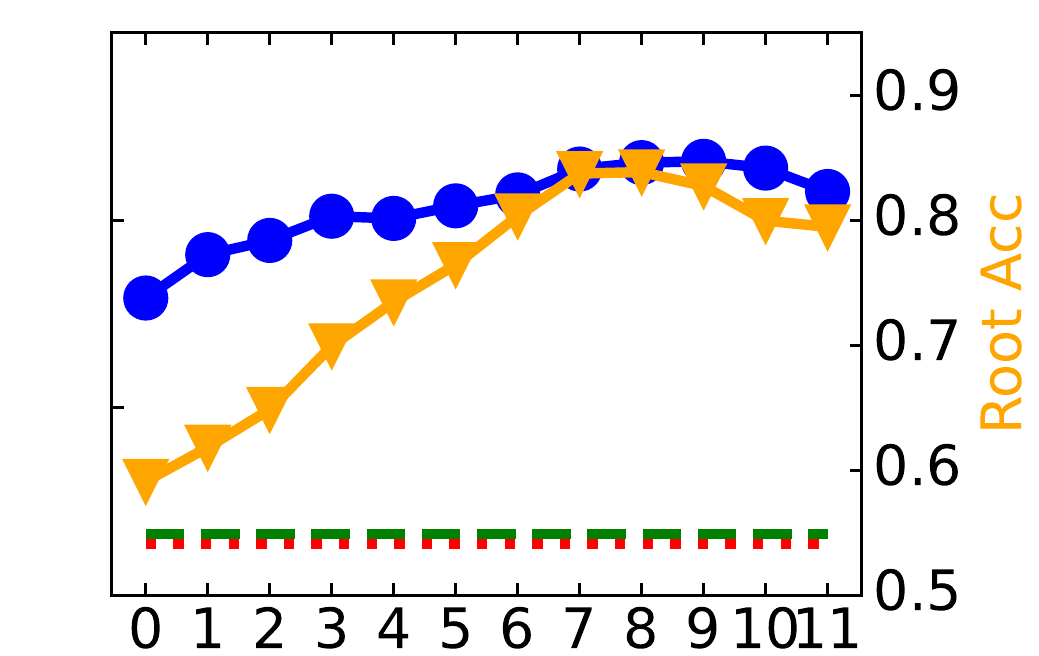}
        \caption{ViLBERT}
        \label{fig:ptb_depth_vil}
    \end{subfigure}
    \caption{Comparison for the depth probe on the PTB3 test set, with textual embeddings.}
    \label{fig:ptb3_depth}
\end{figure*}
\begin{figure*}[t]
    \centering
    \begin{subfigure}[t]{0.9\linewidth}
        \centering
        \includegraphics[scale=0.4]{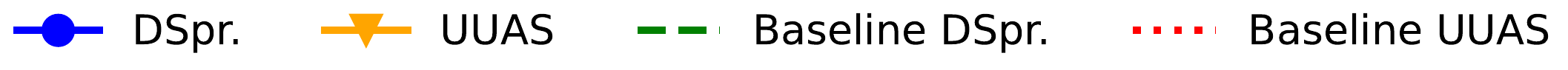}
    \end{subfigure}
    \\
    \begin{subfigure}[t]{0.32\linewidth}
        \centering
        \includegraphics[scale=0.47]{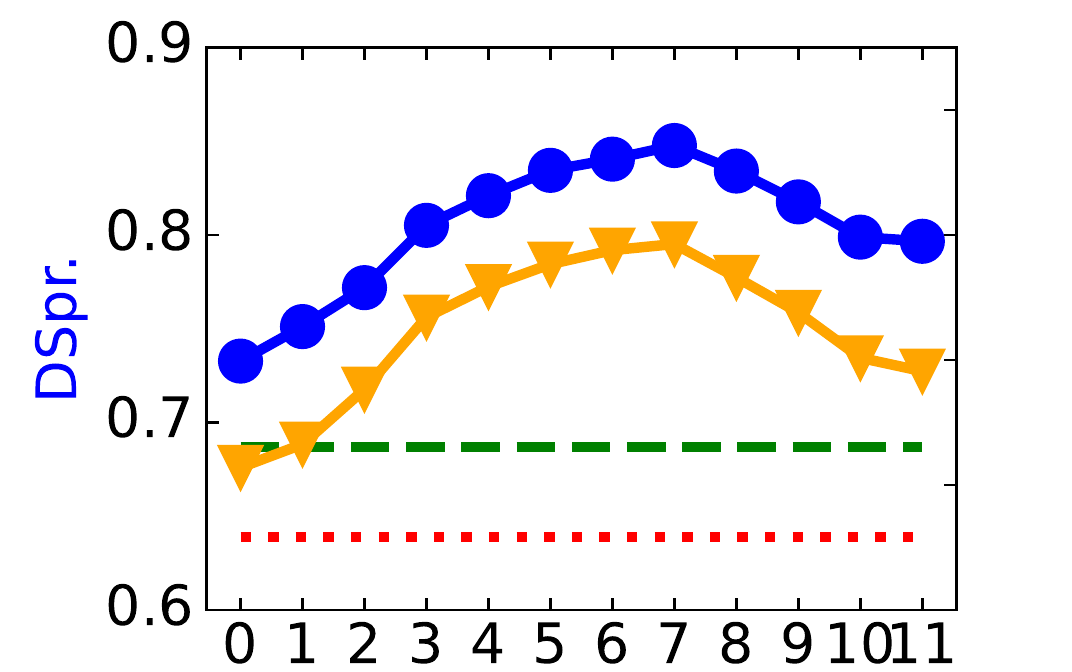}
        \caption{BERT}
        \label{fig:ptb_dist_bert}
    \end{subfigure}
    \begin{subfigure}[t]{0.32\linewidth}
        \centering
        \includegraphics[scale=0.47]{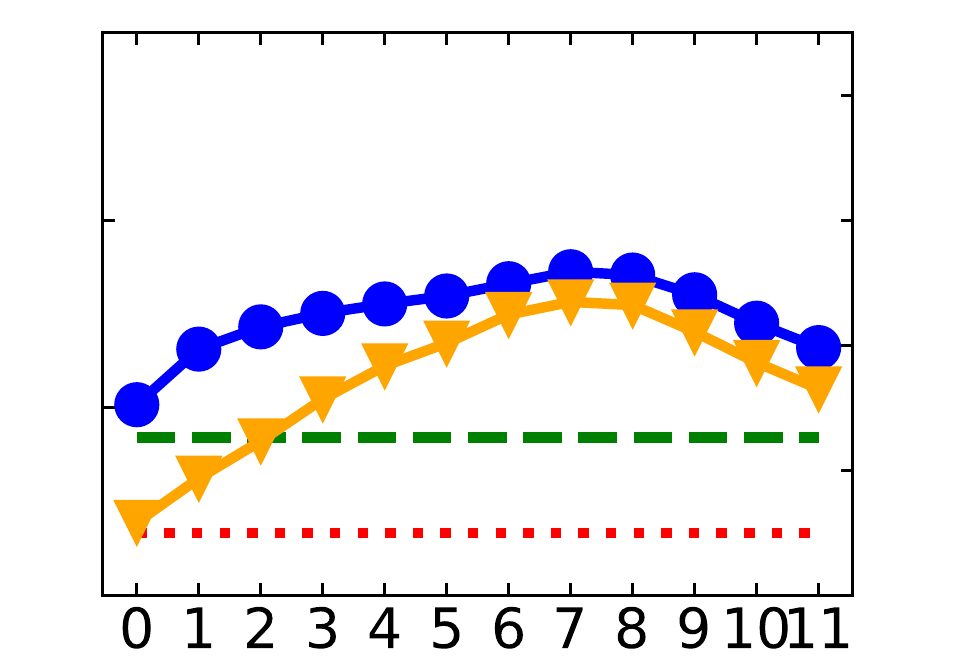}
        \caption{UNITER}
        \label{fig:ptb_dist_unit}
    \end{subfigure}
    \begin{subfigure}[t]{0.32\linewidth}
        \centering
        \includegraphics[scale=0.47]{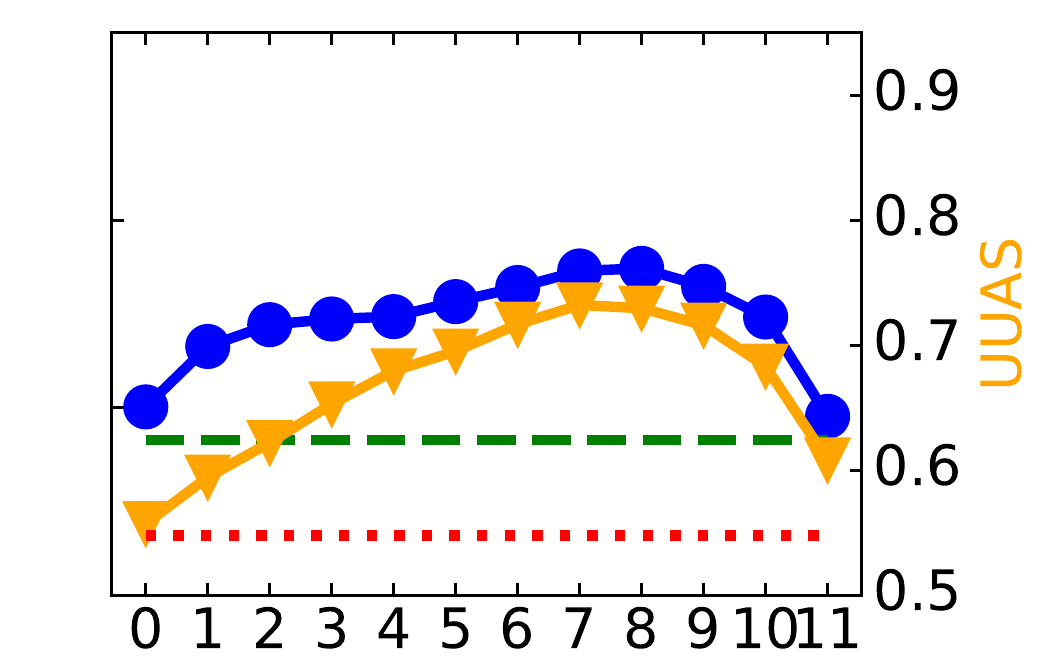}
        \caption{ViLBERT}
        \label{fig:ptb_dist_vil}
    \end{subfigure}
    \caption{Comparison for the distance probe on the PTB3 test set, with textual embeddings.}
\label{fig:ptb3_distance}
\end{figure*}
\begin{figure*}[t]
    \centering
    \begin{subfigure}[t]{0.9\linewidth}
        \centering
        \includegraphics[scale=0.4]{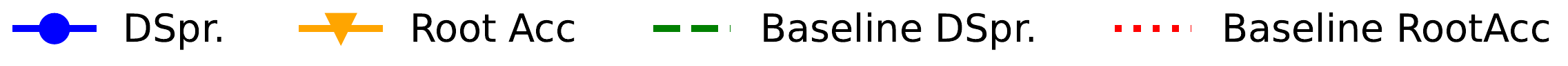}
    \end{subfigure}
    \\
    \begin{subfigure}[t]{0.32\linewidth}
        \centering
        \includegraphics[scale=0.47]{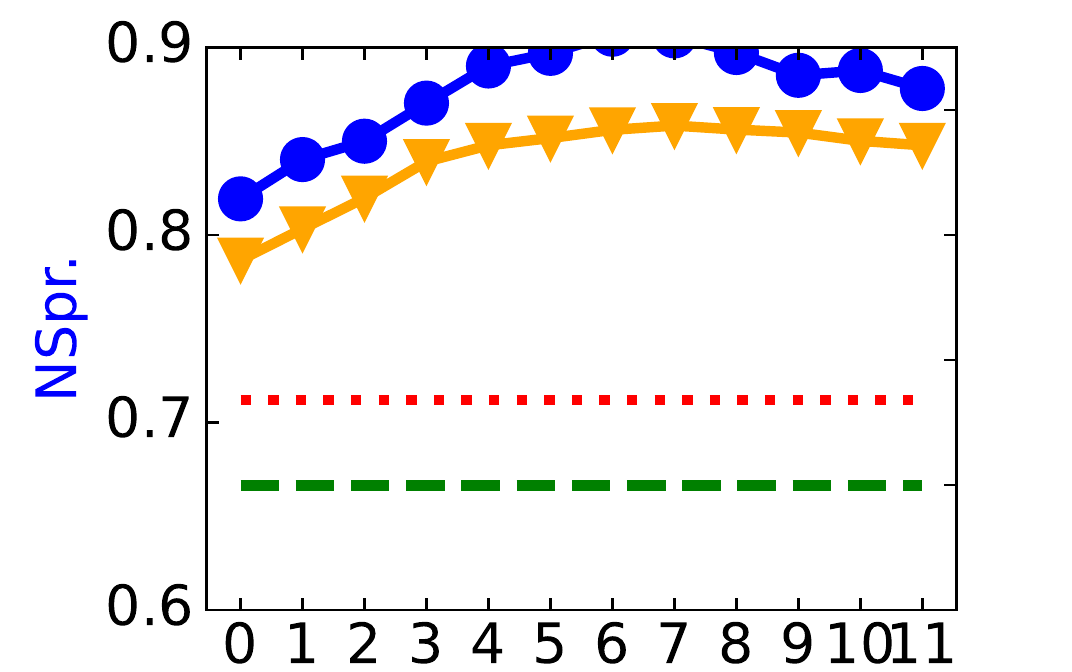}
        \caption{BERT}
        \label{fig:layer_flickr_dep_bert}
    \end{subfigure}
    \begin{subfigure}[t]{0.32\linewidth}
        \centering
        \includegraphics[scale=0.47]{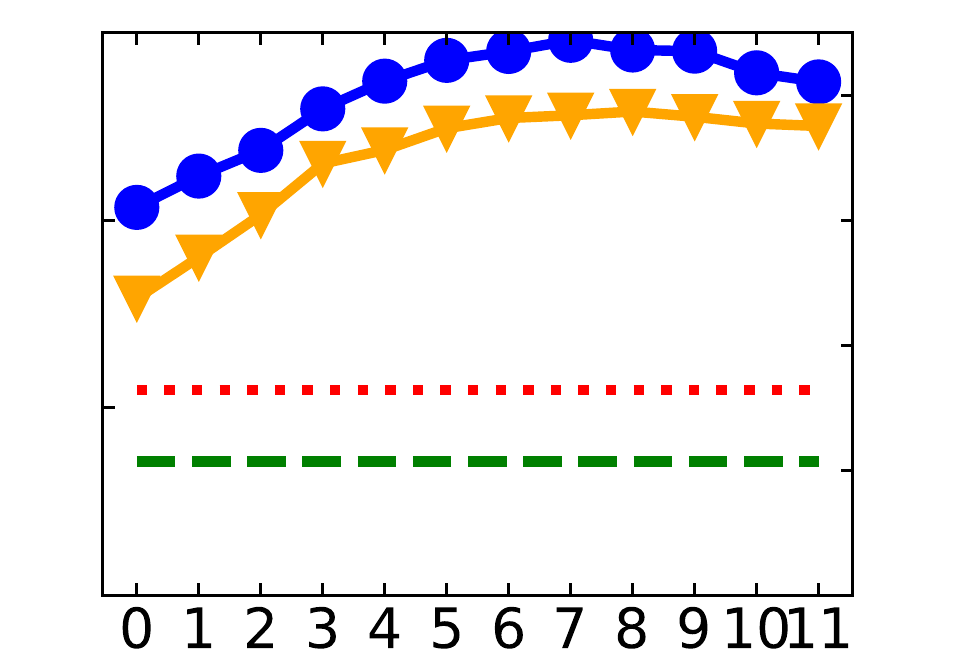}
        \caption{UNITER}
        \label{fig:layer_flickr_dep_unit}
    \end{subfigure}
    \begin{subfigure}[t]{0.32\linewidth}
        \centering
        \includegraphics[scale=0.47]{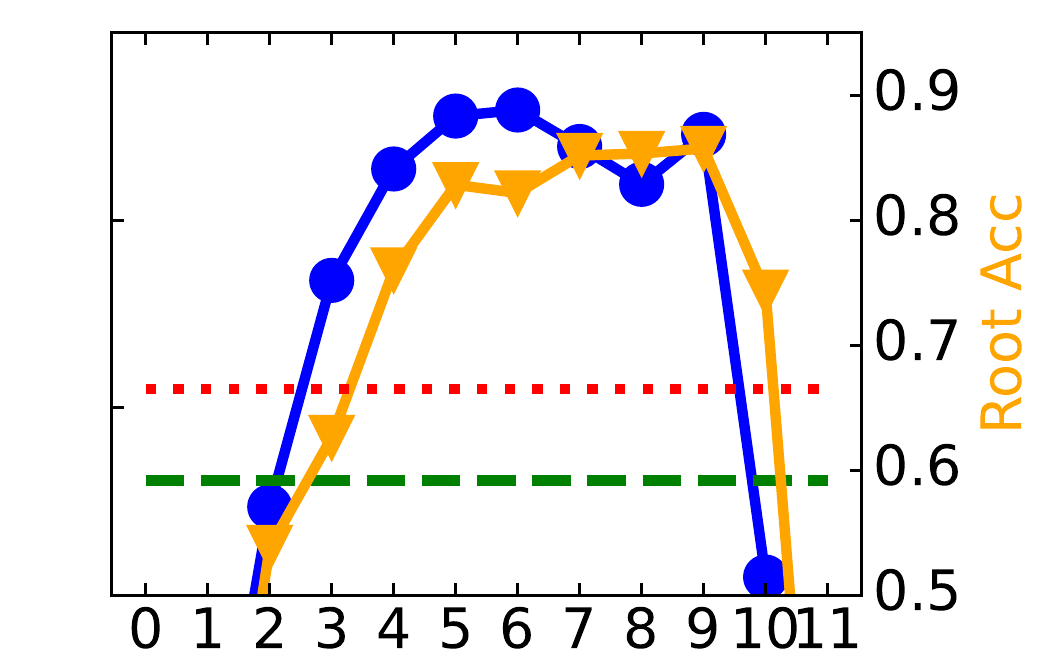}
        \caption{ViLBERT}
        \label{fig:layer_flickr_dep_vil}
    \end{subfigure}
    \caption{Comparison for the depth probe on the Flickr30k test set, with textual embeddings.}
    \label{fig:layer_flickr_dep}
\end{figure*}
\begin{figure*}[t]
    \centering
    \begin{subfigure}[t]{0.9\linewidth}
        \centering
        \includegraphics[scale=0.4]{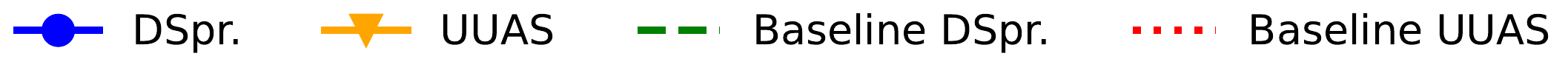}
    \end{subfigure}
    \\
    \begin{subfigure}[t]{0.32\linewidth}
        \centering
        \includegraphics[scale=0.47]{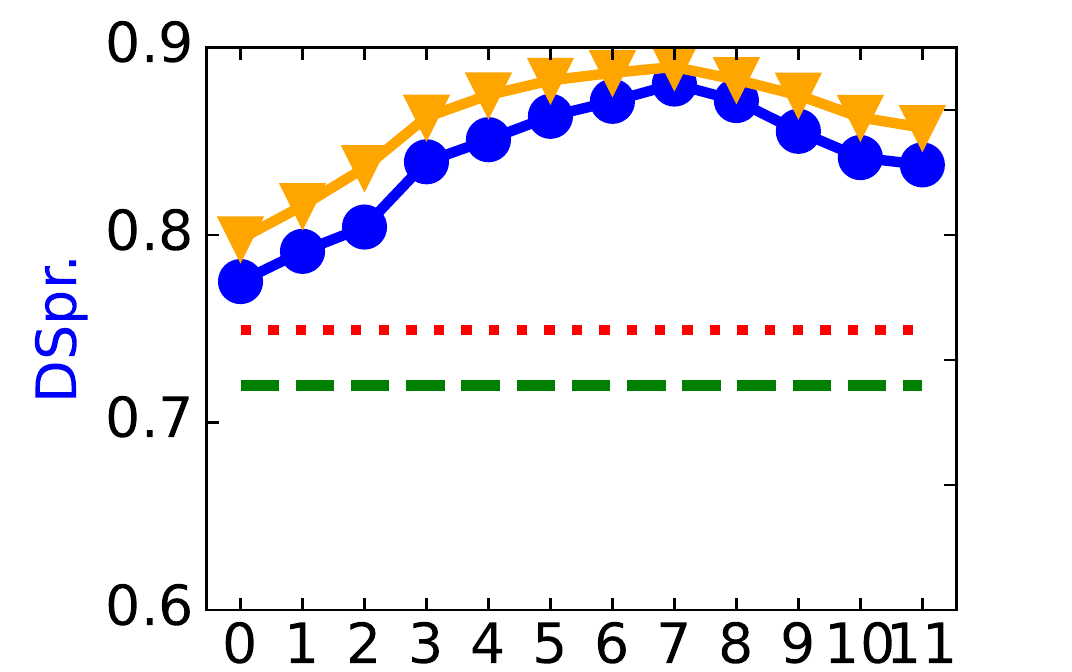}
        \caption{BERT}
        \label{fig:layer_flickr_dist_bert}
    \end{subfigure}
    \begin{subfigure}[t]{0.32\linewidth}
        \centering
        \includegraphics[scale=0.47]{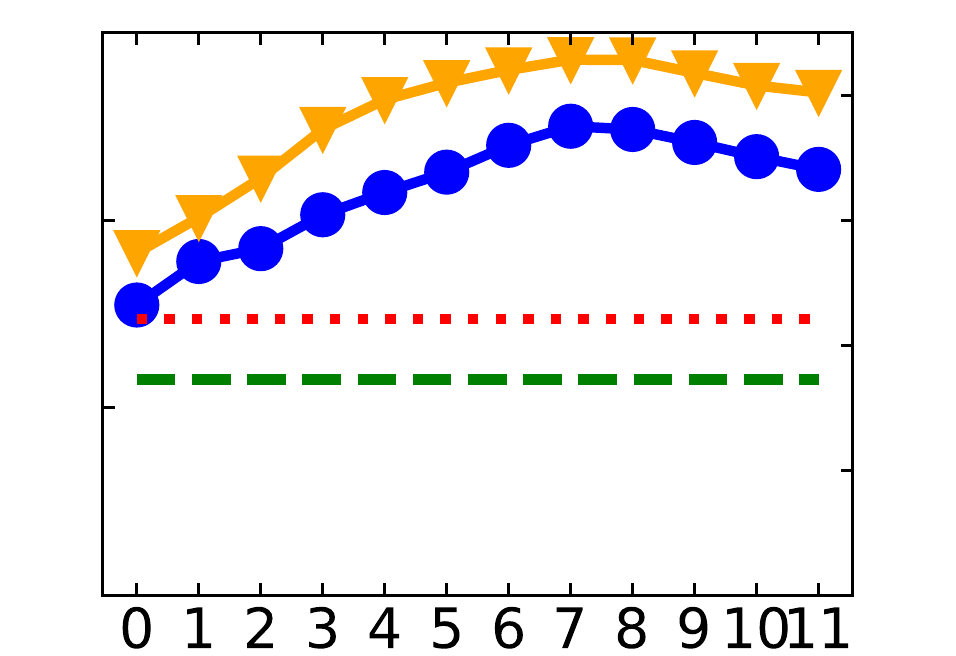}
        \caption{UNITER}
        \label{fig:layer_flickr_dist_unit}
    \end{subfigure}
    \begin{subfigure}[t]{0.32\linewidth}
        \centering
        \includegraphics[scale=0.47]{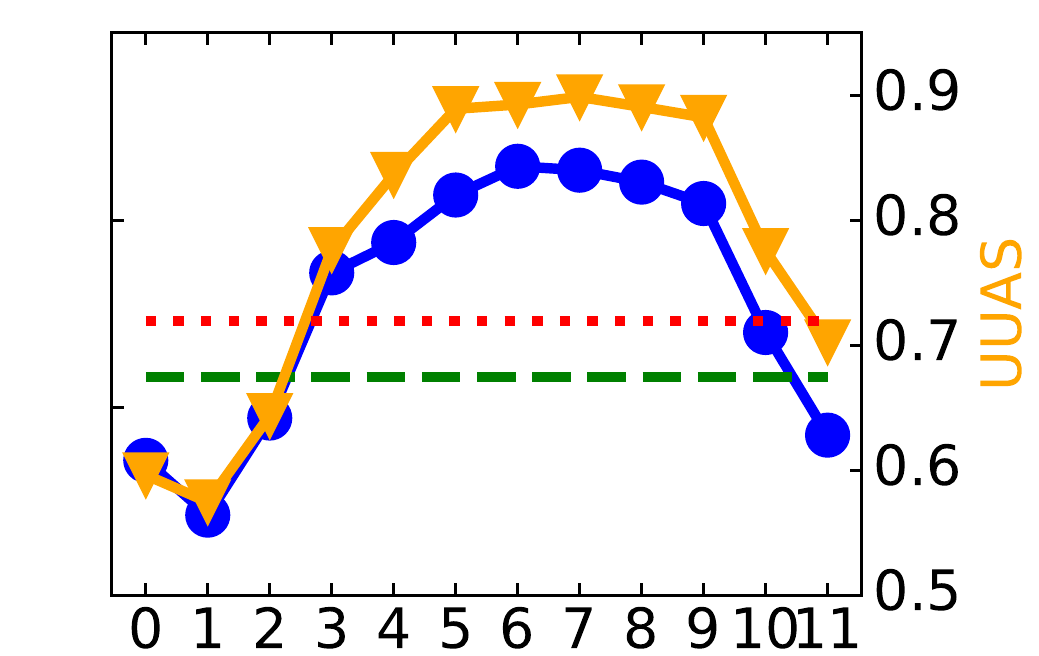}
        \caption{ViLBERT}
        \label{fig:layer_flickr_dist_vil}
    \end{subfigure}
    \caption{Comparison for the distance probe on the Flickr30k test set, with textual embeddings.}
    \label{fig:layer_flickr_dist}
\end{figure*}
\begin{figure*}[t]
    \centering
    \begin{subfigure}[t]{0.9\linewidth}
        \centering
        \includegraphics[scale=0.4]{figures/results/flickr30k/BERT/select_layer_ParseDepthTask_rank128.test.legend.pdf}
    \end{subfigure}
    \\
    \begin{subfigure}[t]{0.32\linewidth}
        \centering
        \includegraphics[scale=0.47]{figures/results/flickr30k/BERT/select_layer_ParseDepthTask_rank128.test.pdf}
        \caption{BERT}
        \label{fig:layer_flickr_dep_bert_just_text}
    \end{subfigure}
    \begin{subfigure}[t]{0.32\linewidth}
        \centering
        \includegraphics[scale=0.47]{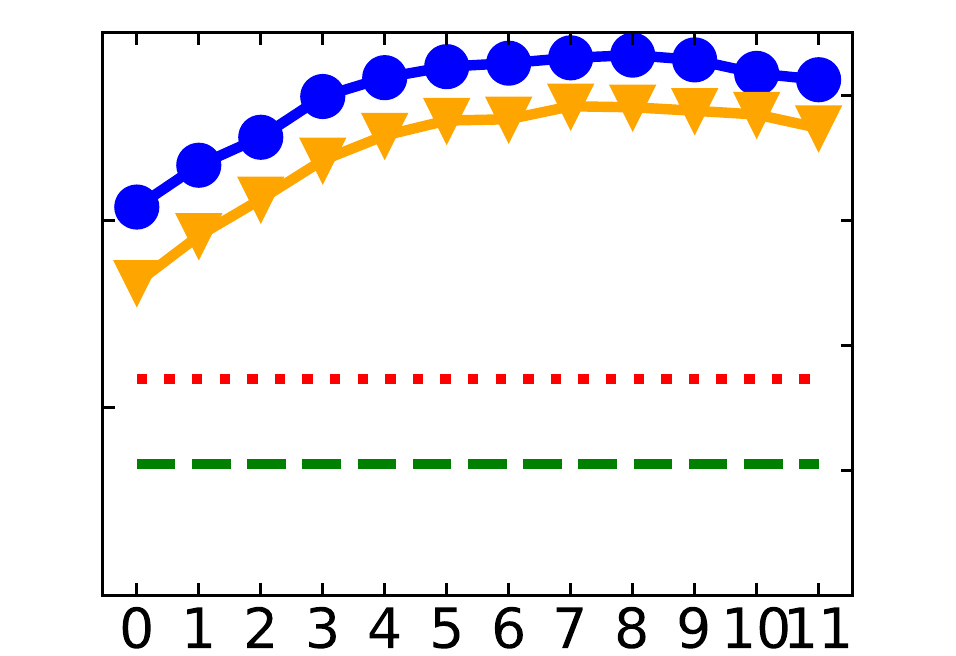}
        \caption{UNITER - only text}
        \label{fig:layer_flickr_dep_unit_just_text}
    \end{subfigure}
    \begin{subfigure}[t]{0.32\linewidth}
        \centering
        \includegraphics[scale=0.47]{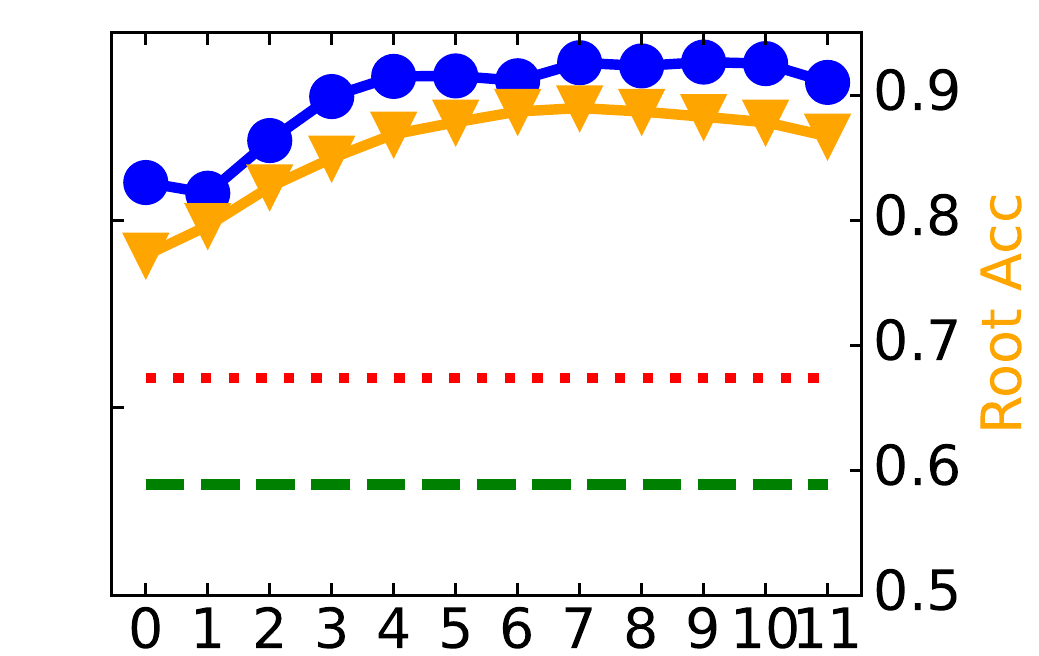}
        \caption{ViLBERT - only text}
        \label{fig:layer_flickr_dep_vil_just_text}
    \end{subfigure}
    \caption{Ablation comparison for the depth probe on the Flickr30k test set while just providing textual embeddings to the multimodal-BERTs.}
    \label{fig:layer_flickr_dep_just_text}
\end{figure*}
\begin{figure*}[t]
    \centering
    \begin{subfigure}[t]{0.9\linewidth}
        \centering
        \includegraphics[scale=0.4]{figures/results/flickr30k/BERT/select_layer_ParseDistanceTask_rank128.test.legend.pdf}
    \end{subfigure}
    \\
    \begin{subfigure}[t]{0.32\linewidth}
        \centering
        \includegraphics[scale=0.47]{figures/results/flickr30k/BERT/select_layer_ParseDistanceTask_rank128.test.pdf}
        \caption{BERT}
        \label{fig:layer_flickr_dist_bert_just_text}
    \end{subfigure}
    \begin{subfigure}[t]{0.32\linewidth}
        \centering
        \includegraphics[scale=0.47]{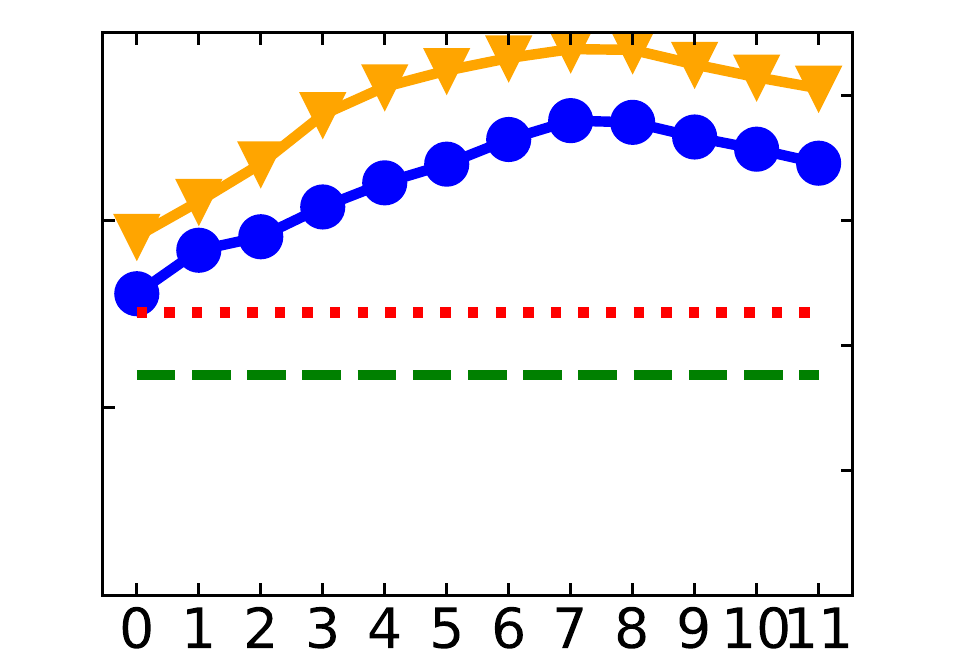}
        \caption{UNITER - only text}
        \label{fig:layer_flickr_dist_unit_just_text}
    \end{subfigure}
    \begin{subfigure}[t]{0.32\linewidth}
        \centering
        \includegraphics[scale=0.47]{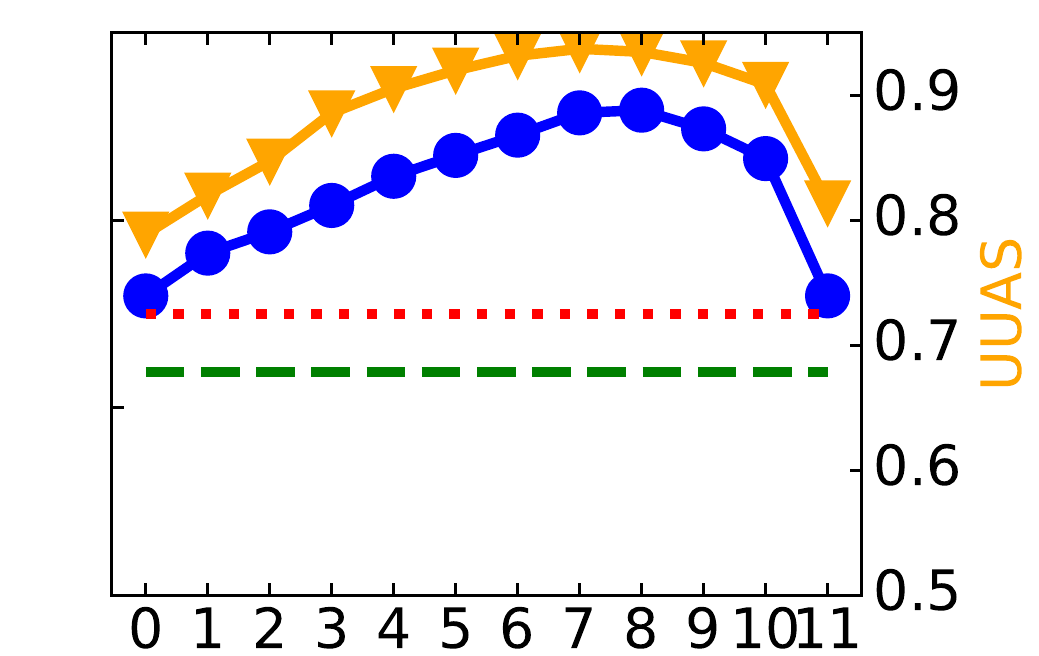}
        \caption{ViLBERT - only text}
        \label{fig:layer_flickr_dist_vil_just_text}
    \end{subfigure}
    \caption{Ablation comparison for the distance probe on the Flickr30k test set while just providing textual embeddings to the multimodal-BERTs.}
    \label{fig:layer_flickr_dist_just_text}
\end{figure*}
\begin{figure*}[t]
\begin{minipage}{.49\textwidth}
\flushleft
    \begin{subfigure}[t]{\linewidth}
        \centering
        \includegraphics[scale=0.4]{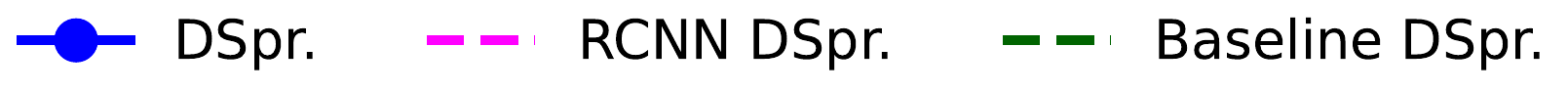}
    \end{subfigure}\\
    \begin{subfigure}[t]{0.48\linewidth}
        \flushleft
        \includegraphics[scale=0.4]{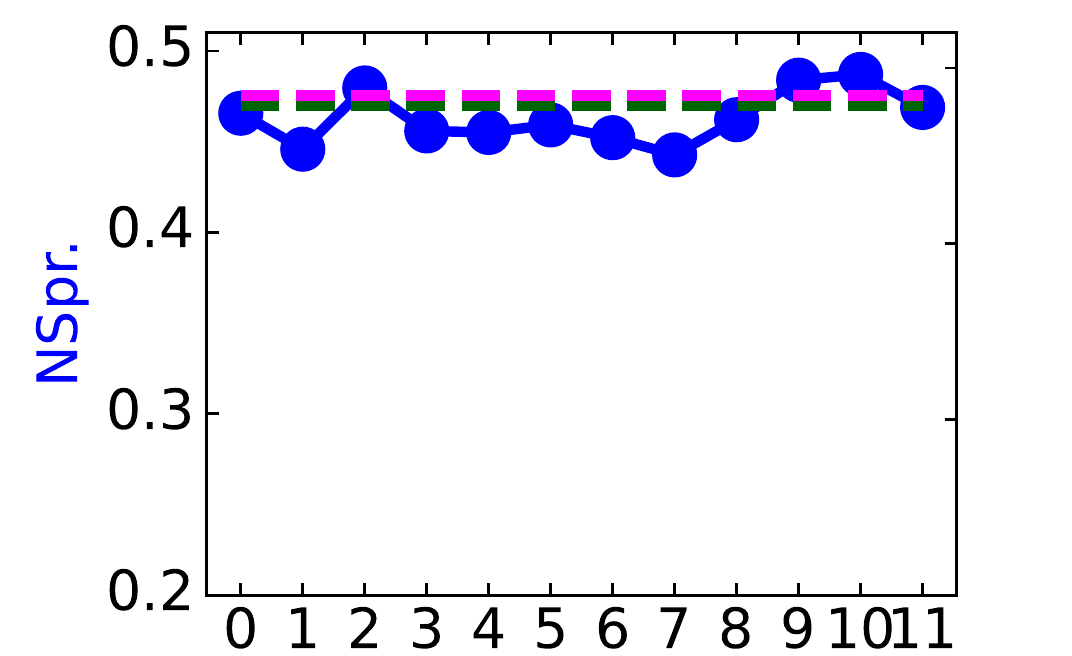}
        \caption{UNITER}
        \label{fig:layer_flickr_visdep_unit}
    \end{subfigure}
    ~
    \begin{subfigure}[t]{0.48\linewidth}
        \flushright
        \includegraphics[scale=0.4]{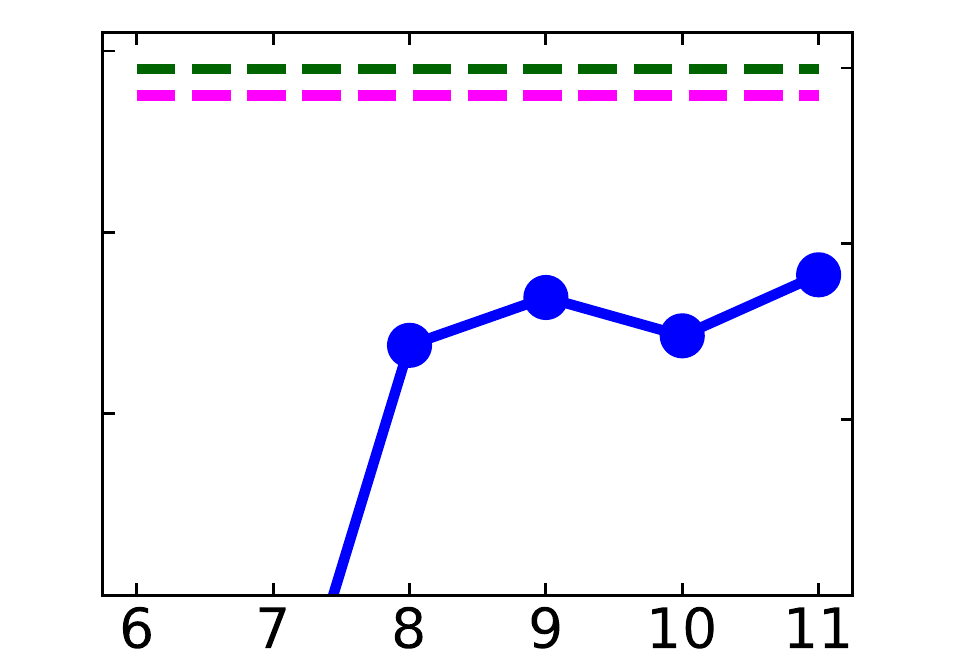}
        \caption{ViLBERT}
        \label{fig:layer_flickr_visdep_vil}
    \end{subfigure}
    \caption{Comparison for the depth probe on the Flickr30k test set, with visual embeddings. Note that the scale is different in this Figure.}
    \label{fig:layer_flickr_visdep}
\end{minipage}
~
\begin{minipage}{.49\textwidth}
    \flushright
    \begin{subfigure}[t]{\linewidth}
        \centering
        \includegraphics[trim={0 10 0 3},clip,scale=0.4]{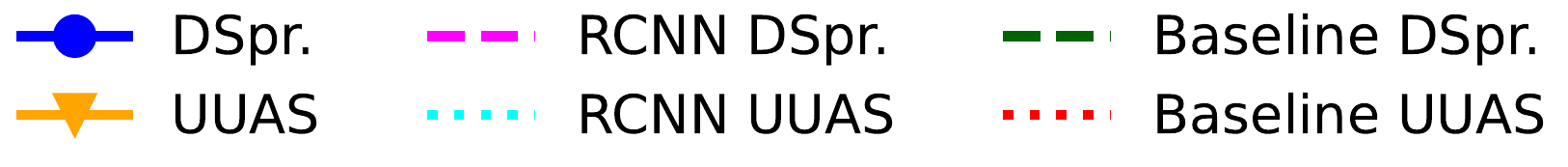}
    \end{subfigure}
    \\
    \begin{subfigure}[t]{0.47\linewidth}
        \flushleft
        \includegraphics[trim={0 0 30 0},clip,scale=0.4]{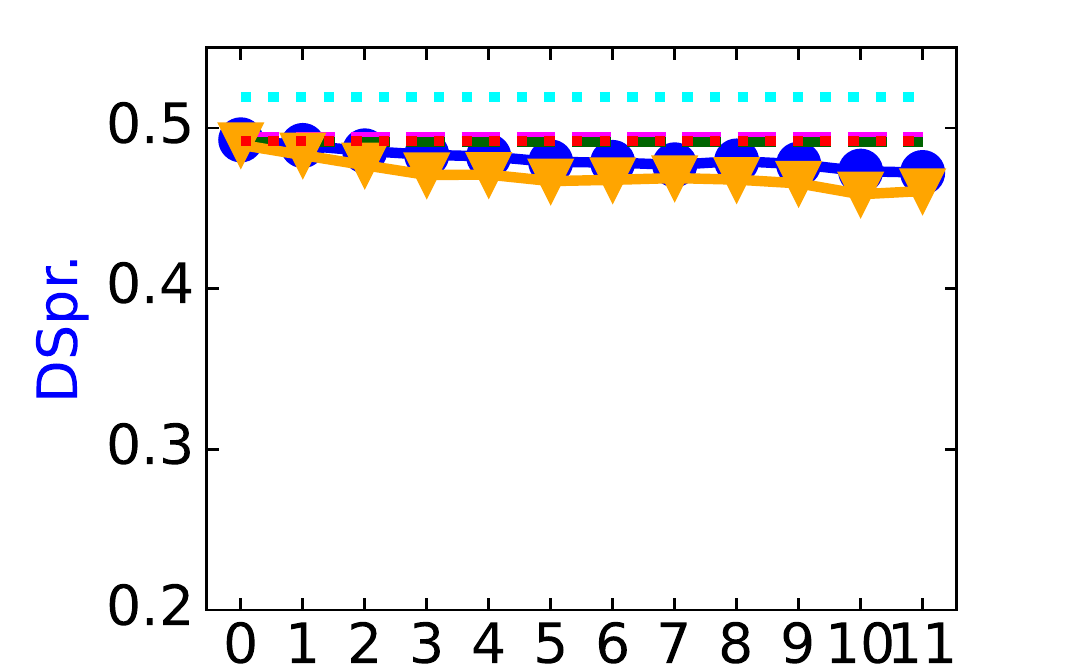}
        \caption{UNITER}
        \label{fig:layer_flickr_visdist_unit}
    \end{subfigure}
    ~
    \begin{subfigure}[t]{0.47\linewidth}
        \flushright
        \includegraphics[trim={30 0 0 0},clip,scale=0.4]{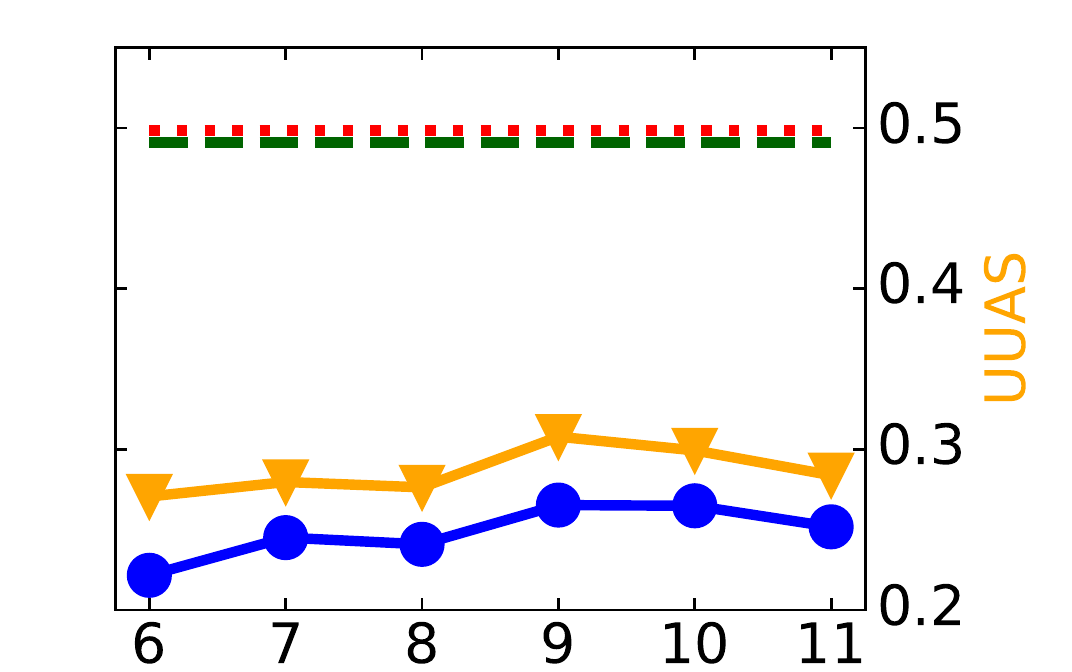}
        \caption{ViLBERT}
        \label{fig:layer_flickr_visdist_vil}
    \end{subfigure}
    \caption{Comparison for the distance probe on the Flickr30k test set, with visual embeddings. Note that the scale is different in this Figure.}
    \label{fig:layer_flickr_visdist}
\end{minipage}
\end{figure*}
The main metric used for both the distance and the depth probes is the Spearman rank coefficient correlation. 
This indicates if the predicted depth vector of the nodes, or the predicted distance matrix of the nodes, correlate with the gold-standard (or silver) depths and distances generated according to the method in Section~\ref{sec:probes}. 
The Spearman correlation is computed for each length sequence separately. 
We take the average over the scores of the lengths between 5 and 50 and call this the Distance Spearman (DSpr.) for the distance probe and the Norm Spearman (NSpr.) for the depth probe.\footnote{Just as done by \citet{hewitt-manning-2019-structural}.} 

For the depth probes, we also use the root accuracy (root\_acc). This computes the accuracy of predicting the root of the sequence. 
This metric is only applicable for the textual embeddings, due to our method of generating the visual tree, where the root is always the full image at the start of the sequence. 

For the distance probe, we make use of the undirected unlabelled attachment score (UUAS). This directly tests how accurate the predicted tree is compared to the ground-truth (or silver) tree by computing the accuracy of predicted connections between nodes in the tree. It does not consider the label for the connection or the direction of the connection \citep{jurafsky2021speech}.

\paragraph{Baseline comparisons}
We design one baseline for the textual data and two for the visual data. For the textual baseline, we use the initial word piece textual embeddings (from either BERT or a multimodal-BERT) before inserting them into the transformer stack. 
We simply refer to it as \textbf{baseline}.

The first visual baseline implements the raw Faster R-CNN features \citep{ren2015faster} of each object region. However, they have a larger dimension than the BERT embeddings. 
We refer to it as \textbf{R-CNN baseline}.
The second baseline uses the visual embeddings before they are fed to the transformer stack. This is a mapping from the Faster R-CNN features to the BERT embedding size. We refer to it as \textbf{baseline}.

\subsection{Hypotheses}\label{sec:hypotheses}
First, we want to determine the probe rank of the linear transformation used on the textual or the visual embeddings. 
Based on results by \citet{hewitt-manning-2019-structural}, we set the probe rank for BERT to 128. We run a comparison with several probe ranks on UNITER and ViLBERT to find the optimal setting for the textual and visual embeddings. The results are shown and discussed in Appendix~\ref{app:tune_rank}. We use a rank of 128 for all our following experiments. 

\paragraph{RQ 1} 
The multimodal-BERT models are pre-trained on language data. We assume that the resulting embeddings integrate structural grammatical knowledge and hypothesize that this knowledge will not be forgotten during multimodal training.
 
To determine if training on multimodal data affects the quality of predicting the dependency tree when trained solely with textual data, we train the probes with BERT and both multimodal-BERTs and evaluate on the PTB3 dataset \citep{marcus1999treebank}. 

\paragraph{Sub-RQ 1.1}
We expect that more interaction between the regions and the text will have a stronger impact. Some dependency attachments that are hard to predict might require visual knowledge. 
Next to the effect on the linguistic knowledge, we also want to discover if the multimodal data helps the multimodal-BERTs in learning structural knowledge. 
We run the probes on Flickr30k dataset \citep{young2014flickr30k} with the textual embeddings for all our models. Furthermore, we compare these to the difference in scores on the PTB3 dataset \citep{marcus1999treebank}. 

\paragraph{RQ 2}
The Multimodal-BERTs learn highly contextualized embeddings. Therefore, we hypothesize that a model should be able to discover important interactions between object regions in the image. 
To see if the model has learned to encode the scene tree in the visual region embeddings, we run the probes on the Flickr30k dataset \citep{young2014flickr30k} with the visual embeddings. 
Furthermore, to see if the scene tree is learned mainly through joint interaction with the textual embeddings, we compare the scores between the single-stream model UNITER (with many cross-modal interactions) and the dual-stream model ViLBERT (with limited cross-modal interactions).

\section{Results and Discussion}\label{sec:results}
This discussion is based on the results from the test split. The results on the validation split (see Appendix~\ref{sec:val_results}), lead to the same observations.

\paragraph{RQ 1: Do the textual embeddings trained with a multimodal-BERT retain their structural knowledge?}
To answer RQ 1, we report the results for both structural probes on the PTB3 dataset. Here we only use the textual embeddings, since no visual features are available. The results for the depth probe are in Figure~\ref{fig:ptb3_depth}, and for the distance probe in Figure~\ref{fig:ptb3_distance}. 

The results of both multimodal-BERTs (Figures~\ref{fig:ptb_depth_vil} and \ref{fig:ptb_dist_vil} for ViLBERT and Figures~\ref{fig:ptb_depth_unit} and \ref{fig:ptb_dist_unit} for UNITER) in terms of NSpr. and Root Acc are very comparable showing similar curves and scores. 
For both, the seventh layer is the best performing one. The shape of the curves across the layers is similar to those for the BERT model in Figures~\ref{fig:ptb_depth_bert} and \ref{fig:ptb_dist_bert}. 
However, the scores of the multimodal-BERTs drop significantly. 
While the multimodal-BERTs were initialized with weights from BERT, they were trained longer on additional multimodal data with a different multimodal objective. This shows that the multimodal training hampers the storing of grammatical structural knowledge in the resulting embeddings. 

\paragraph{Sub-RQ 1.1: To what extent does the joint training in a multimodal-BERT influence the structures learned in the textual embeddings?}
For this experiment, we compare the effect of having visual features present when using the structural probes on the textual embeddings. We run the probes on Flickr30k. The results for the depth probe are in Figure~\ref{fig:layer_flickr_dep}, and for the distance probe in Figure~\ref{fig:layer_flickr_dist}. 

First, we see that for all models (BERT and multimodal-BERTs) the scores increase compared to the results on the PTB3 dataset (see discussion of RQ 1), but still follow a similar trend across the layers. 
The latter is most likely due to the complexity of the sentences and language of the PTB3 dataset, which is simpler for the captions. 
For ViLBERT, there is a drop in performance for the earlier layers. We believe this is caused by the early stopping method firing early with these settings. Another explanation is that it is more difficult for the dual-stream model to use the additional parameters.

BERT outperforms the multimodal-BERTs on PTB3, however, this is not the case on Flickr30k. 
For the depth probe (Figure~\ref{fig:layer_flickr_dep}) and the UUAS metric on the distance probe (Figure~\ref{fig:layer_flickr_dist}), the results obtained on these two datasets are almost equal.
This can be due to the additional pretraining of the multimodal-BERTs on similar captioning sentences.
Another explanation is that, during such pretraining, the models learned to store relevant information in the visual embeddings. 

We run an additional experiment where we use the pretrained multimodal-BERT, but while probing we only provide the sentence to the model, and mask out the image. The results for the depth probe are in Figure~\ref{fig:layer_flickr_dep_just_text}, and for the distance probe in Figure~\ref{fig:layer_flickr_dist_just_text}. Here we can see that the results are almost identical to when we provide the model with the visual embeddings. This indicates that the model does not have any benefit from the visual data when predicting the structures for textual embeddings, and it seems that the model uses the extra parameters of the vision layers to store knowledge about the text.

\paragraph{RQ 2: Do the visual embeddings trained with a multimodal-BERT learn to encode a scene tree?}
We aim to find the layer with the most structural knowledge learned when applied to multimodal data. See the results in Figures~\ref{fig:layer_flickr_visdep} and \ref{fig:layer_flickr_visdist}.

Regarding the results for the depth probe (Figure~\ref{fig:layer_flickr_visdep}), the scores between layers fluctuate inconsistently. The scores do improve slightly over the baselines, indicating that the multimodal-BERT encodes some knowledge of depth in the layers. 

With regard to the distance probe (Figure~\ref{fig:layer_flickr_visdist}), the trend in the curves across the layers indicate that this is a type of knowledge that can be learned for the regions. 
The multimodal-BERTs seem to disregard scene trees. There is a strong downward trend across the layers. 
Furthermore, all the scores are much lower than the baseline and the R-CNN baseline scores.
This lack of learning of the scene tree can be caused by the chosen training objective of the multimodal-BERTs.
These objectives require an abstract type of information, where only basic features are needed to predict the masked items.

For the distance probe, there is a noticeable difference between the single-stream  (Figure~\ref{fig:rank_flickr_visdist_unit}) and the dual-stream (Figure~\ref{fig:rank_flickr_visdist_vil}) models, where single stream models benefit from the multimodal interactions to retain structural knowledge. 
For UNITER, the scores in the first layers are very close to the baseline, showing that the single stream interaction benefits the memorizing of the scene tree structure.

\section{Conclusion and Future Work}
We made a first attempt at investigating whether the current Multimodal-BERT models encode structural grammatical knowledge in their textual embeddings, in a similar way as text-only BERT models encode this knowledge. 
Furthermore, we were the first to investigate the existence of encoded structural compositional knowledge of the object regions in image embeddings. 
For this purpose, we created a novel scene tree structure that is mapped from the textual dependency tree of the paired caption. 
We discovered that the multimodal-BERTs encode less structural grammatical knowledge than BERT. However, with image features present, it is still possible to achieve similar results. The cause for this requires more research. 

While tree depths from the scene tree are not natively present in the features, we found that this could be a potential method of finding connections and distances between regions, already decently predicted with the Faster R-CNN features. 
The Multimodal-BERT models are currently trained with an objective 
that does not enforce the learning or storing of these types of structural information. Hence we assume that the models learn to encode more abstract knowledge in their features. 

Our work opens possibilities to further research on scene trees as a joint representation of object compositions in an image and the grammatical structure of its caption. 
Furthermore, we recommend investigating the training of multimodal-BERTs with objectives that enforce the encoding of structural knowledge.

\section*{Acknowledgments}
We would like to thank Desmond Elliott, Djam\'e Seddah, and Liesbeth Allein for feedback on the paper.
Victor Milewski and Marie-Francine Moens were funded by the European Research Council (ERC) Advanced Grant CALCULUS (grant agreement No. 788506). 
Miryam de Lhoneux was funded by the Swedish Research Council (grant 2020-00437).
\bibliographystyle{acl_natbib}
\bibliography{bib}

\appendix
\begin{figure*}[htb]
\begin{minipage}{.49\textwidth}
    \flushleft
    \begin{subfigure}[t]{\linewidth}
        \centering
        \includegraphics[scale=0.4]{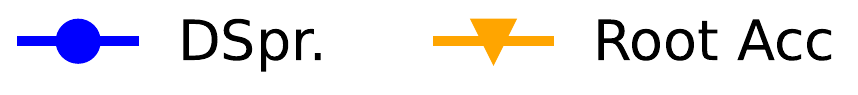}
    \end{subfigure}
    \\
    \begin{subfigure}[t]{0.47\linewidth}
        \flushleft
        \includegraphics[trim={0 0 5 0},clip,scale=0.4]{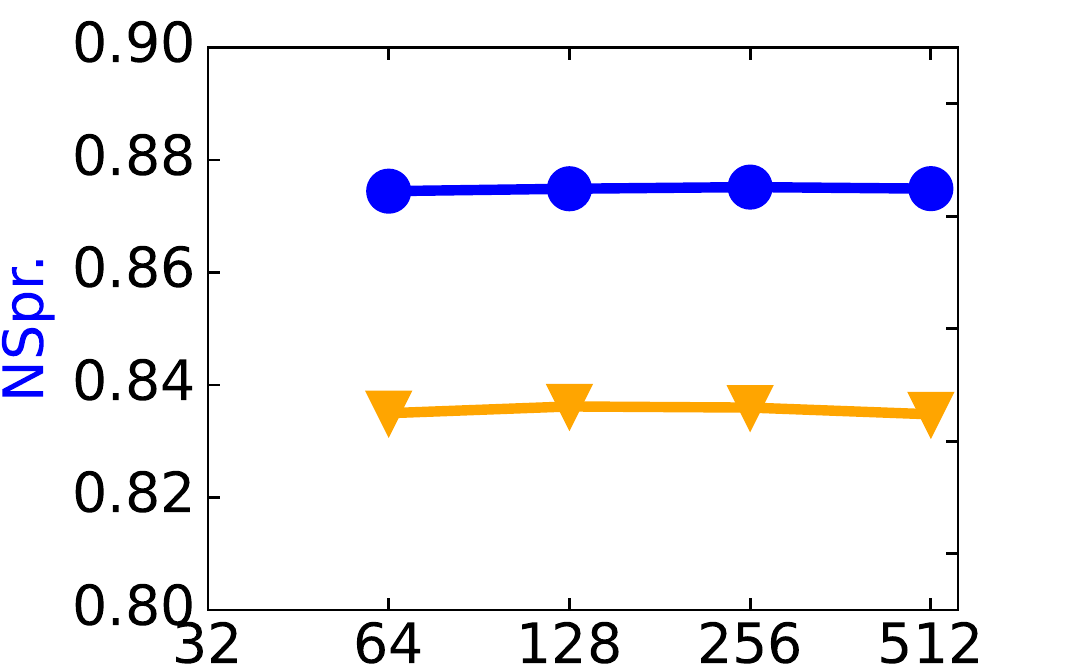}
        \caption{UNITER}
        \label{fig:rank_flickr_dep_unit}
    \end{subfigure}
    ~
    \begin{subfigure}[t]{0.47\linewidth}
        \flushright
        \includegraphics[trim={30 0 0 -4},clip,scale=0.4]{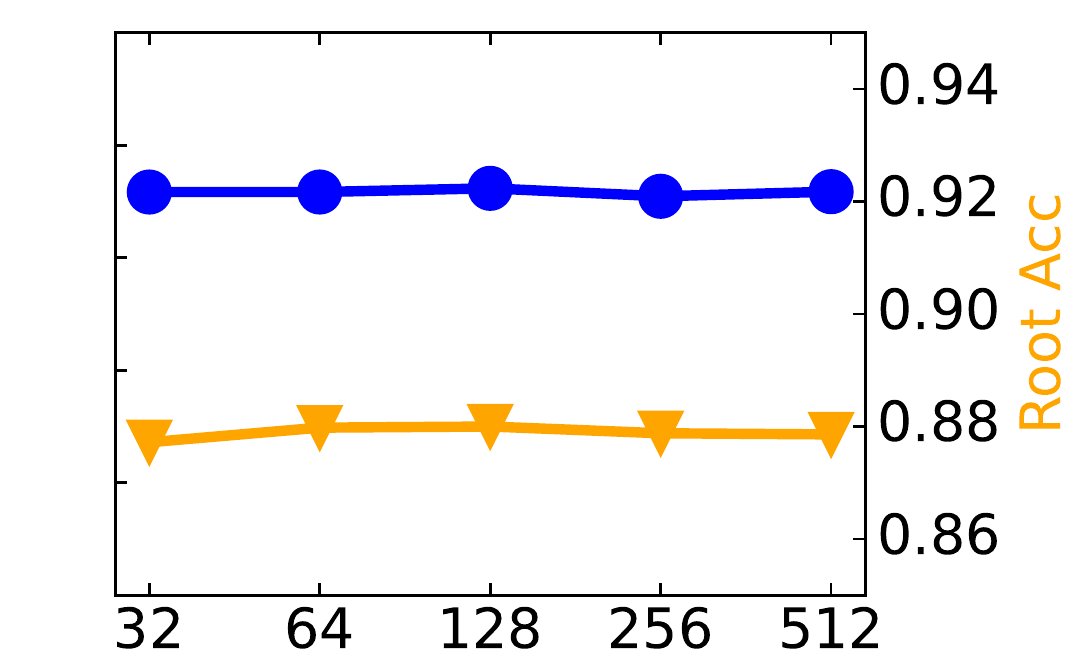}
        \caption{ViLBERT}
        \label{fig:rank_flickr_dep_vil}
    \end{subfigure}
    \caption{Tuning the depth probe rank on the textual embeddings.}
    \label{fig:rank_flickr_dep}
\end{minipage}
\begin{minipage}{.49\textwidth}
\centering
    \begin{subfigure}[t]{\linewidth}
        \centering
        \includegraphics[scale=0.4]{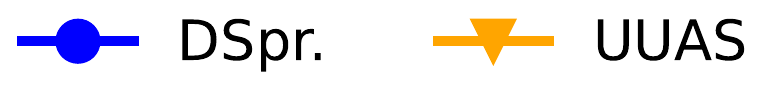}
    \end{subfigure}
    \\
    \begin{subfigure}[t]{0.47\linewidth}
        \flushleft
        \includegraphics[scale=0.4]{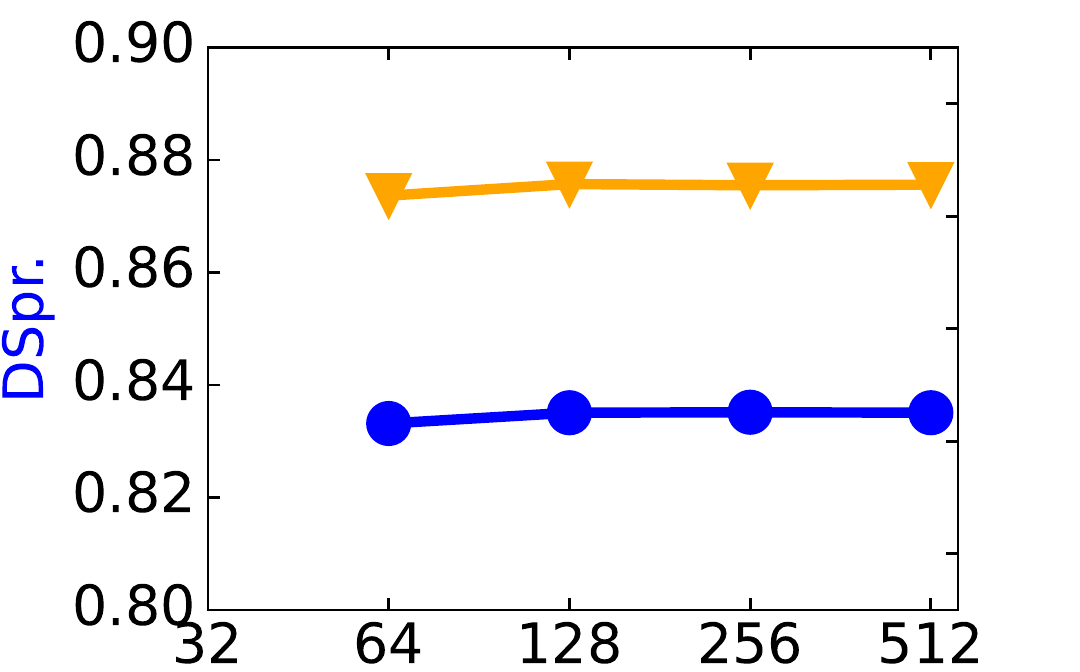}
        \caption{UNITER}
        \label{fig:rank_flickr_dist_unit}
    \end{subfigure}
    \hspace{5pt}
    \begin{subfigure}[t]{0.47\linewidth}
        \flushright
        \includegraphics[trim={30 0 0 -5},clip,scale=0.4]{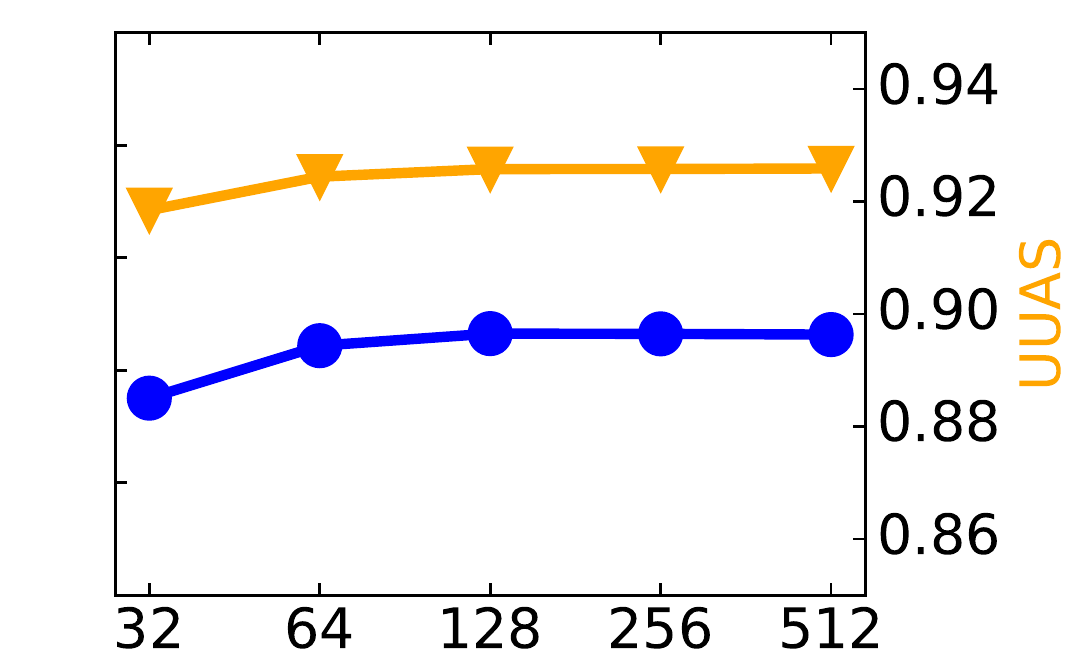}
        \caption{ViLBERT}
        \label{fig:rank_flickr_dist_vil}
    \end{subfigure}
    \caption{Tuning the distance probe rank on the textual embeddings.}
    \label{fig:rank_flickr_dist}
\end{minipage}
\end{figure*}
\begin{figure*}[htb]
\begin{minipage}{.49\textwidth}
    \flushleft
    \begin{subfigure}[t]{\linewidth}
        \centering
        \includegraphics[scale=0.4]{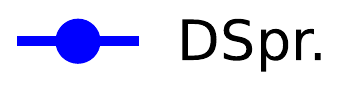}
    \end{subfigure}
    \\
    \begin{subfigure}[t]{0.47\linewidth}
        \flushleft
        \includegraphics[scale=0.4]{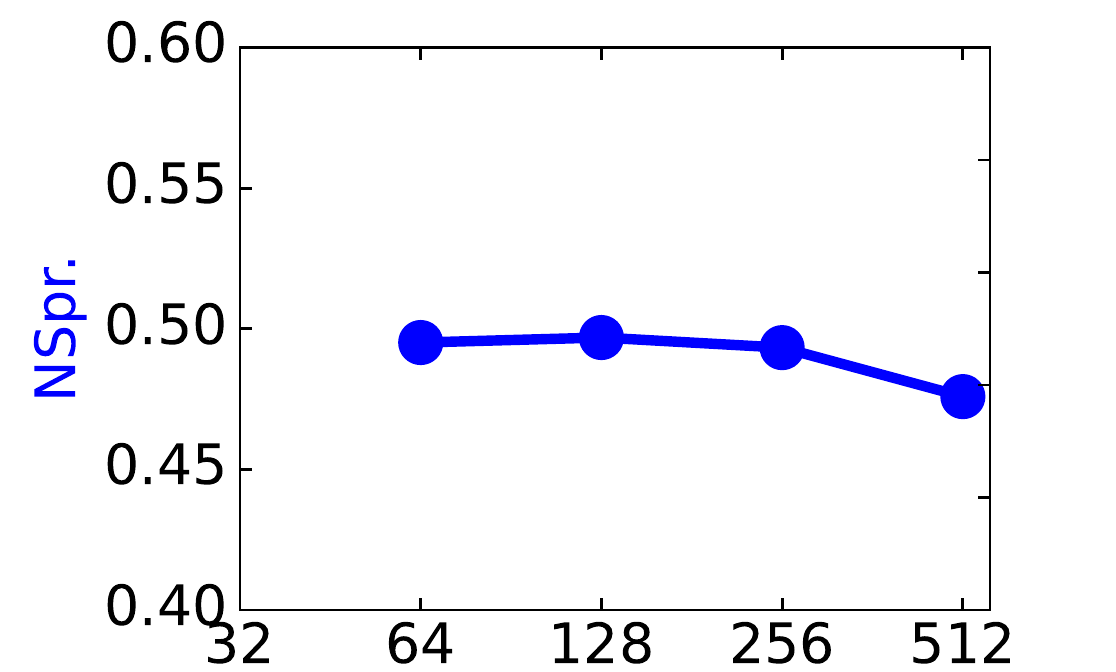}
        \caption{UNITER}
        \label{fig:rank_flickr_visdep_unit}
    \end{subfigure}
    \hspace{6pt}
    \begin{subfigure}[t]{0.47\linewidth}
        \flushright
        \includegraphics[trim={5 0 0 -5},clip,scale=0.4]{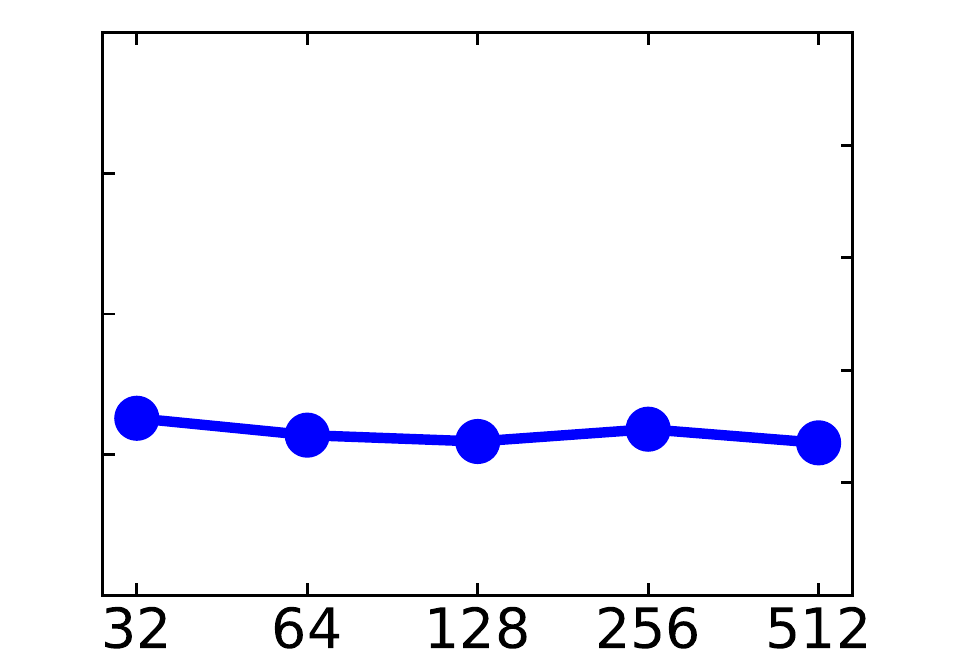}
        \caption{ViLBERT}
        \label{fig:rank_flickr_visdep_vil}
    \end{subfigure}
    \caption{Tuning the depth probe rank on the visual embeddings.}
    \label{fig:rank_flickr_visdep}
\end{minipage}
\begin{minipage}{.49\textwidth}
    \flushleft
    \begin{subfigure}[t]{\linewidth}
        \centering
        \includegraphics[scale=0.4]{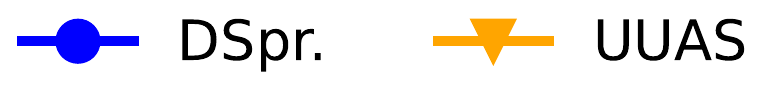}
    \end{subfigure}
    \\
    \begin{subfigure}[t]{0.47\linewidth}
        \flushleft
        \includegraphics[scale=0.4]{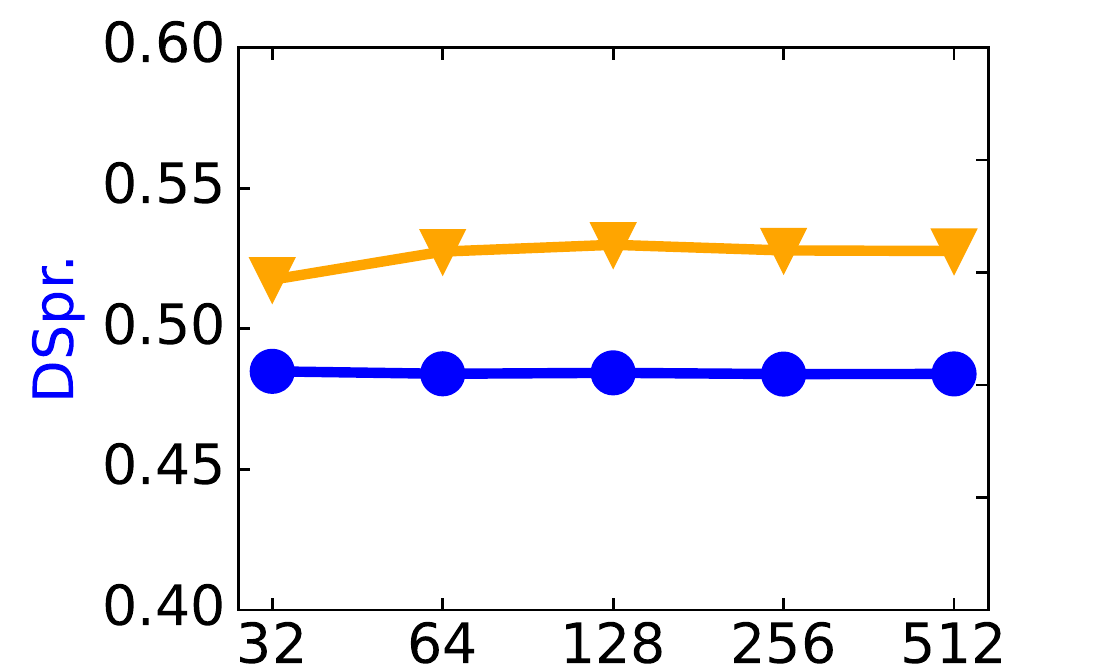}
        \caption{UNITER}
        \label{fig:rank_flickr_visdist_unit}
    \end{subfigure}
    \hspace{5.5pt}
    \begin{subfigure}[t]{0.47\linewidth}
        \flushright
        \includegraphics[trim={30 0 0 0},clip,scale=0.4]{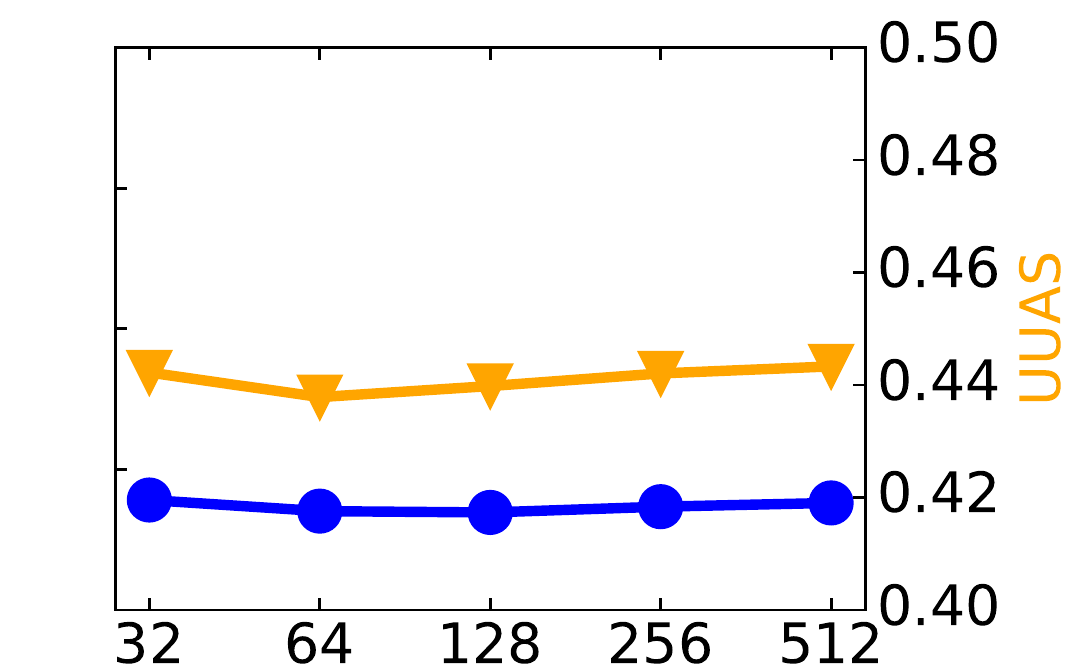}
        \caption{ViLBERT}
        \label{fig:rank_flickr_visdist_vil}
    \end{subfigure}
    \caption{Tuning the distance probe rank on the visual embeddings.}
    \label{fig:rank_flickr_visdist}
\end{minipage}
\end{figure*}
\section{Tuning Probe Rank}\label{app:tune_rank}
To find the dimensionality needed for the multimodal-BERTS, we made a comparison between several probes. The results for the textual embeddings are in Figures~\ref{fig:rank_flickr_dep} and \ref{fig:rank_flickr_dist}. Here we see that the probe rank does not have any significant effect of changing the performance of the models. Therefore, we decided it is best to follow the optimal rank found for the BERT model: 128.

The results for the visual embeddings are in Figures~\ref{fig:rank_flickr_visdep} and \ref{fig:rank_flickr_visdist}. 
Here we also see only very small changes. Therefore, we also keep the probe rank at 128 for the visual features. 

\section{Results on Validation Split}\label{sec:val_results}
The same graphs as for our experiments discussed in Section~\ref{sec:results} using the validation set instead of the test set. 
The graphs created for the test set are very similar to those the validation set. The results lead to an identical conclusion. 
One difference is the performance of the ViLBERT model. On the textual features, the score for earlier layers is again comparable with the other models. This indicates that the early stopping indead fired to early. 

Furthermore, ViLBERT is less capable to predict the scene trees, which confirms the hypothesis that inter-modal interaction is needed to learn the structural knowledge that is implicitly present in the image and its captions.

\begin{figure*}[htb]
    \centering
    \begin{subfigure}[t]{0.9\linewidth}
        \centering
        \includegraphics[scale=0.4]{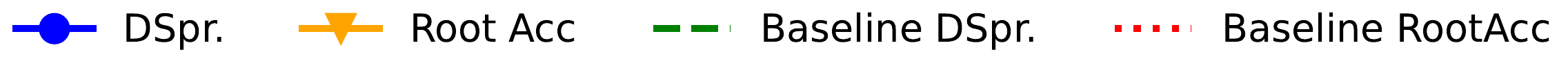}
    \end{subfigure}
    \\
    \begin{subfigure}[t]{0.32\linewidth}
        \centering
        \includegraphics[scale=0.48]{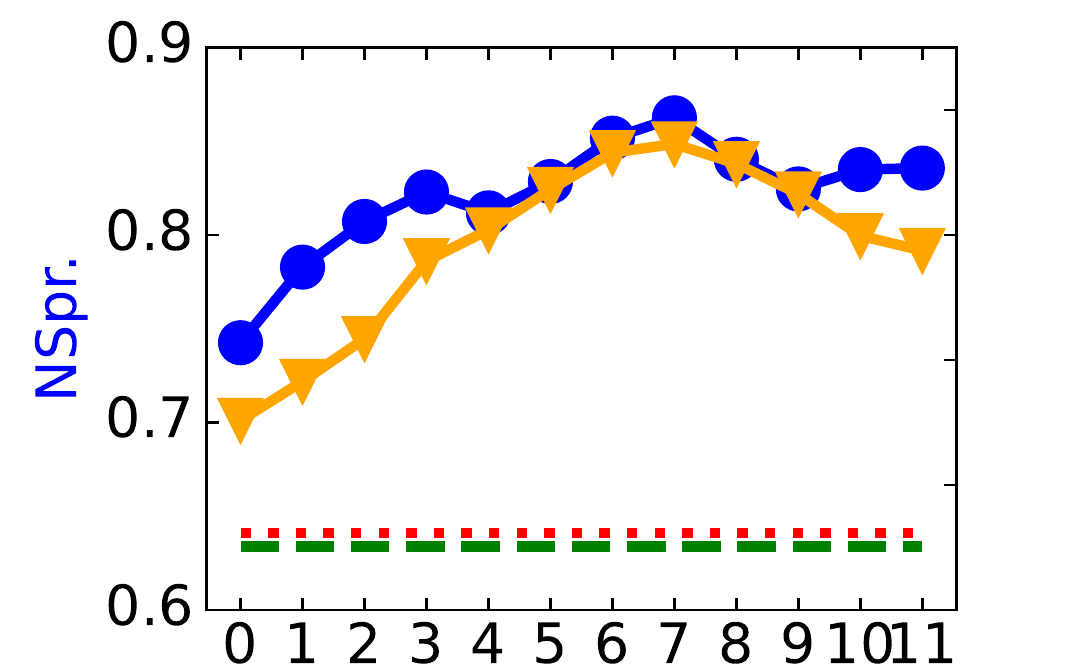}
        \caption{BERT}
        \label{fig:ptb_depth_bert_val}
    \end{subfigure}
    \begin{subfigure}[t]{0.32\linewidth}
        \centering
        \includegraphics[scale=0.48]{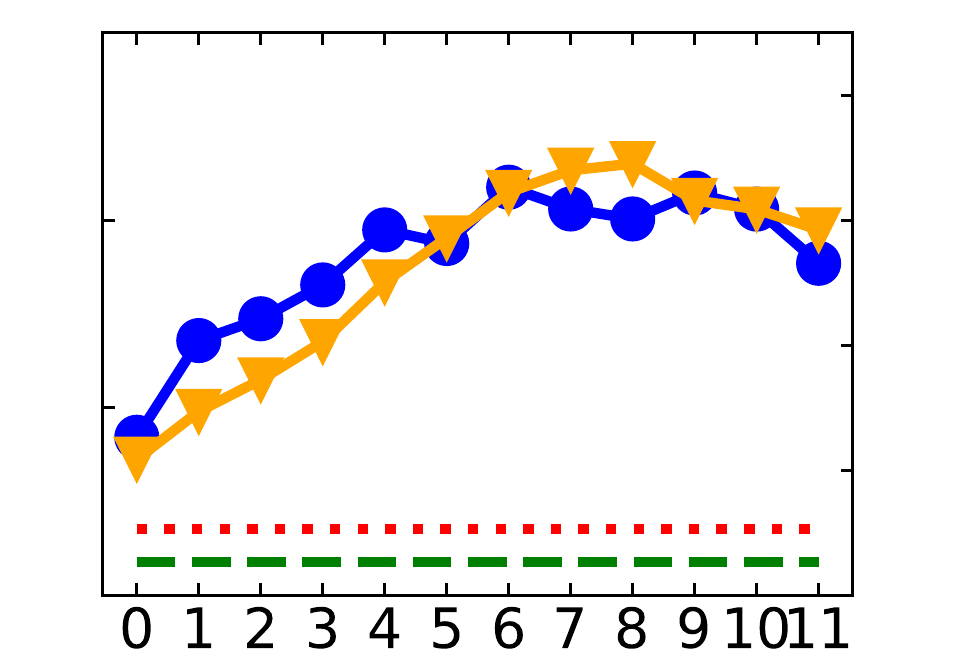}
        \caption{UNITER}
        \label{fig:ptb_depth_unit_val}
    \end{subfigure}
    \begin{subfigure}[t]{0.32\linewidth}
        \centering
        \includegraphics[scale=0.48]{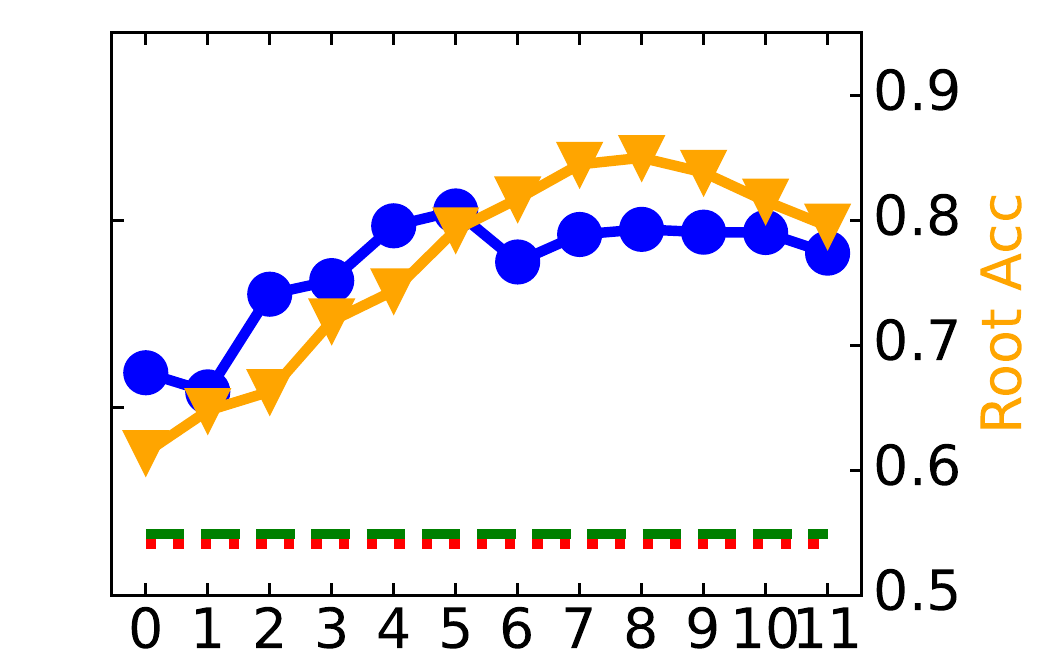}
        \caption{ViLBERT}
        \label{fig:ptb_depth_vil_val}
    \end{subfigure}
    \caption{Comparison for the depth probe on the PTB3 validation set, with textual embeddings.}
    \label{fig:ptb3_depth_val}
\end{figure*}
\begin{figure*}[htb]
    \centering
    \begin{subfigure}[t]{0.9\linewidth}
        \centering
        \includegraphics[scale=0.4]{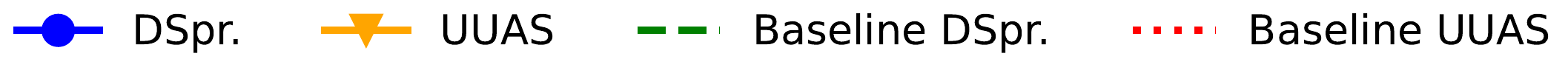}
    \end{subfigure}
    \\
    \begin{subfigure}[t]{0.32\linewidth}
        \centering
        \includegraphics[scale=0.47]{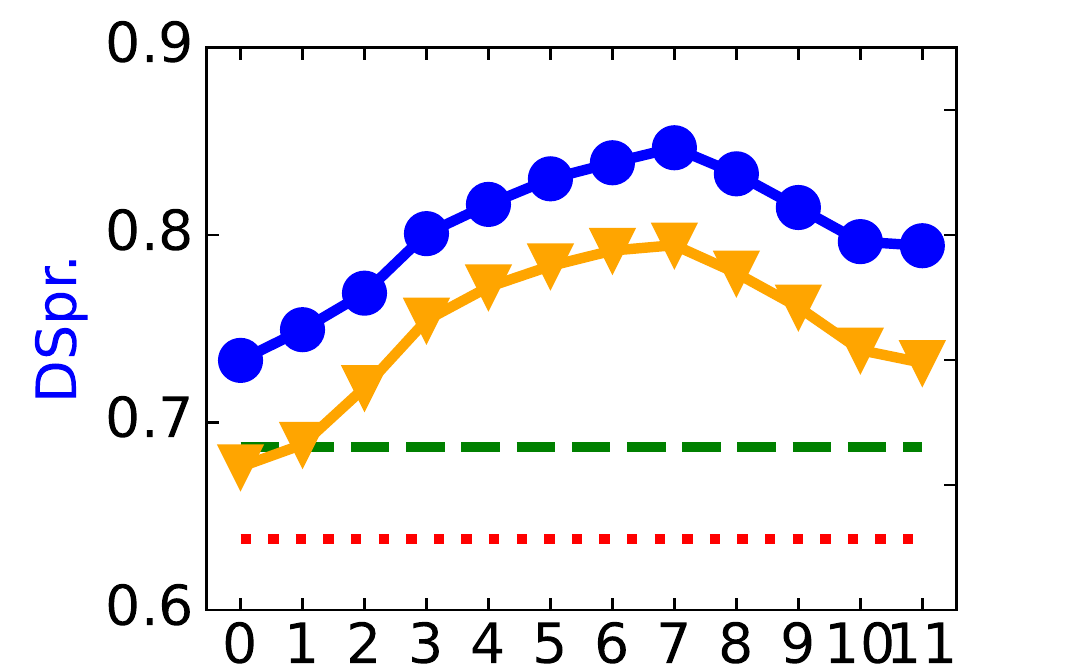}
        \caption{BERT}
        \label{fig:ptb_dist_bert_val}
    \end{subfigure}
    \begin{subfigure}[t]{0.32\linewidth}
        \centering
        \includegraphics[scale=0.47]{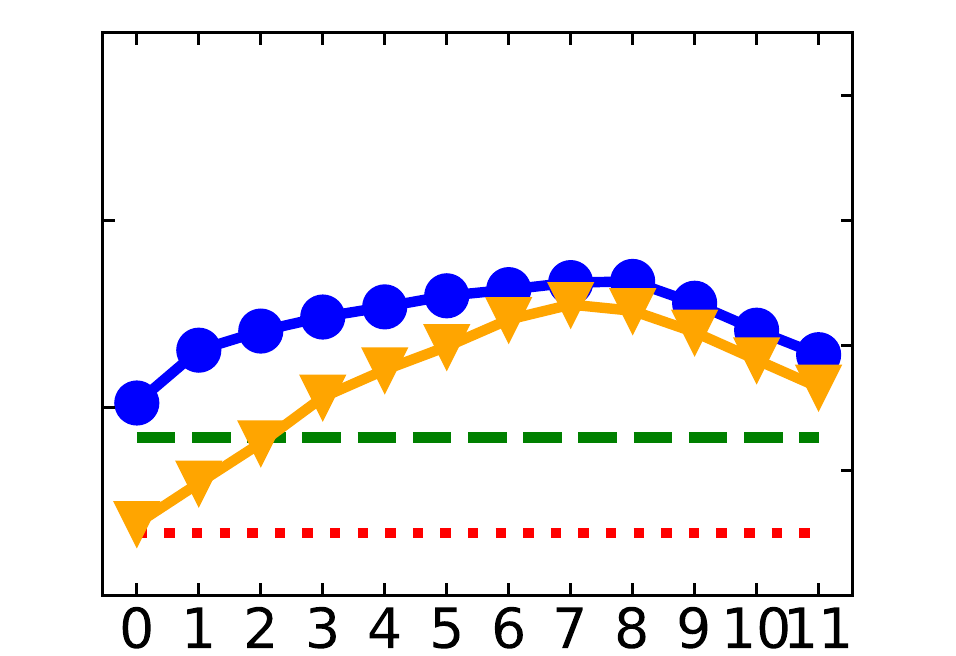}
        \caption{UNITER}
        \label{fig:ptb_dist_unit_val}
    \end{subfigure}
    \begin{subfigure}[t]{0.32\linewidth}
        \centering
        \includegraphics[scale=0.47]{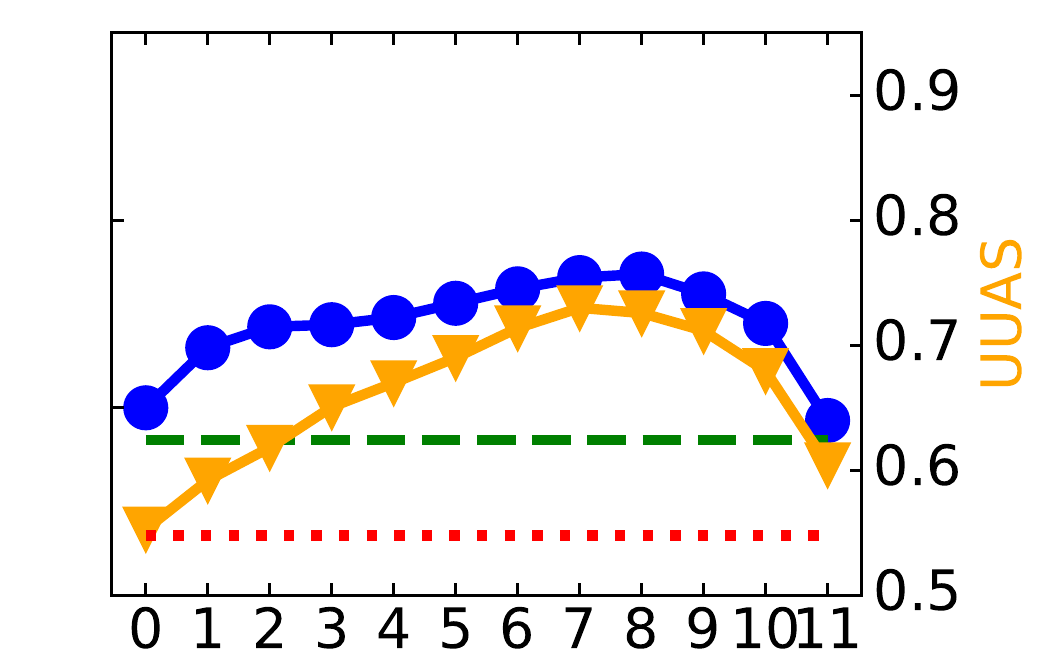}
        \caption{ViLBERT}
        \label{fig:ptb_dist_vil_val}
    \end{subfigure}
    \caption{Comparison for the distance probe on the PTB3 validation set, with textual embeddings.}
\label{fig:ptb3_distance_test}
\end{figure*}
\begin{figure*}[htb]
    \centering
    \begin{subfigure}[t]{0.9\linewidth}
        \centering
        \includegraphics[scale=0.4]{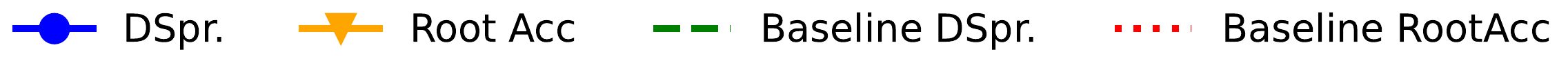}
    \end{subfigure}
    \\
    \begin{subfigure}[t]{0.32\linewidth}
        \centering
        \includegraphics[scale=0.47]{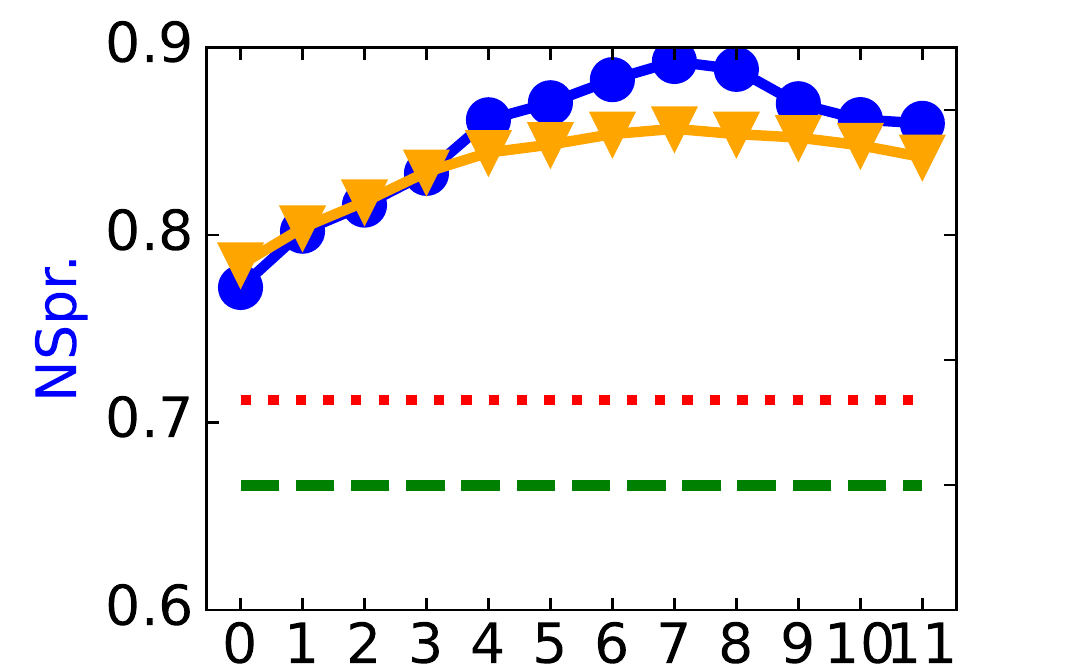}
        \caption{BERT}
        \label{fig:layer_flickr_dep_bert_val}
    \end{subfigure}
    \begin{subfigure}[t]{0.32\linewidth}
        \centering
        \includegraphics[scale=0.47]{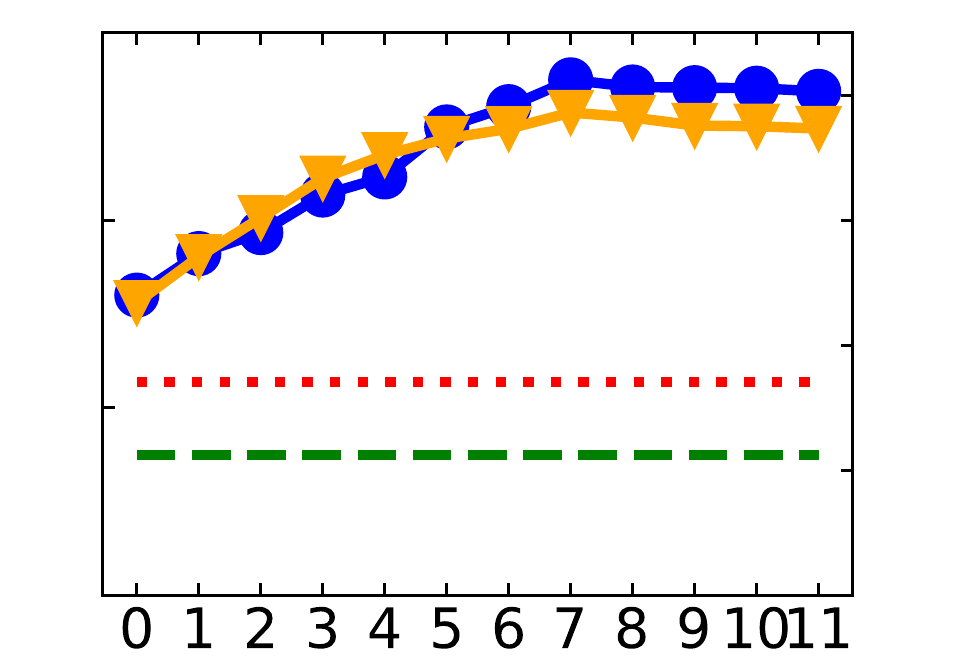}
        \caption{UNITER}
        \label{fig:layer_flickr_dep_unit_val}
    \end{subfigure}
    \begin{subfigure}[t]{0.32\linewidth}
        \centering
        \includegraphics[scale=0.47]{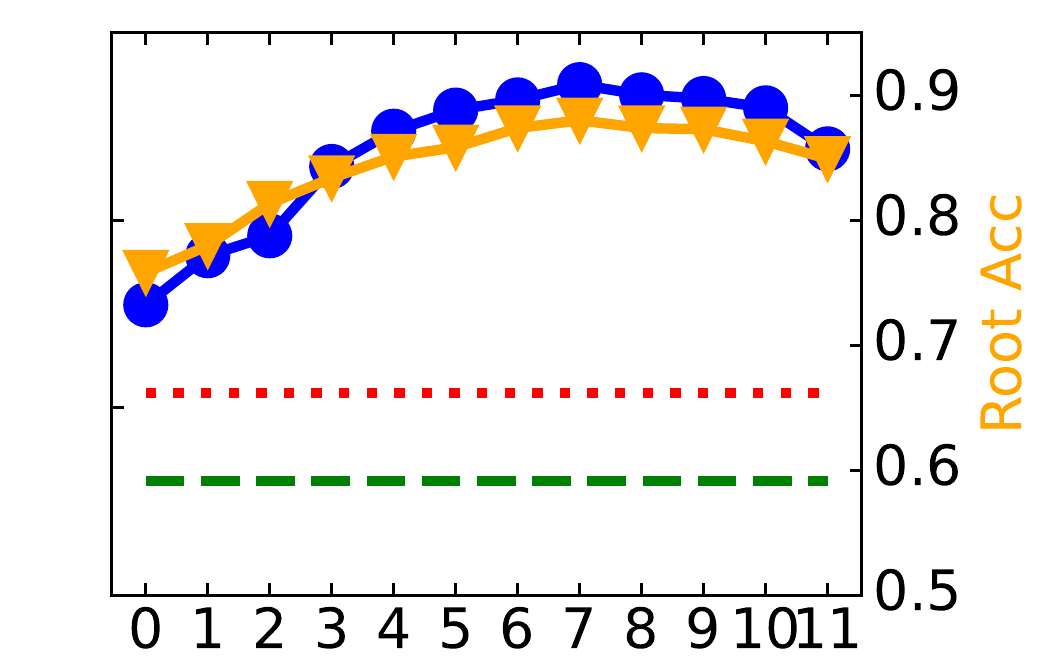}
        \caption{ViLBERT}
        \label{fig:layer_flickr_dep_vil_val}
    \end{subfigure}
    \caption{Comparison for the depth probe on the Flickr30k validation set, with textual embeddings.}
    \label{fig:layer_flickr_dep_val}
\end{figure*}
\begin{figure*}[htb]
    \centering
    \begin{subfigure}[t]{0.9\linewidth}
        \centering
        \includegraphics[scale=0.4]{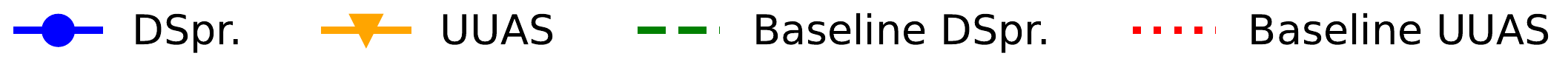}
    \end{subfigure}
    \\
    \begin{subfigure}[t]{0.32\linewidth}
        \centering
        \includegraphics[scale=0.47]{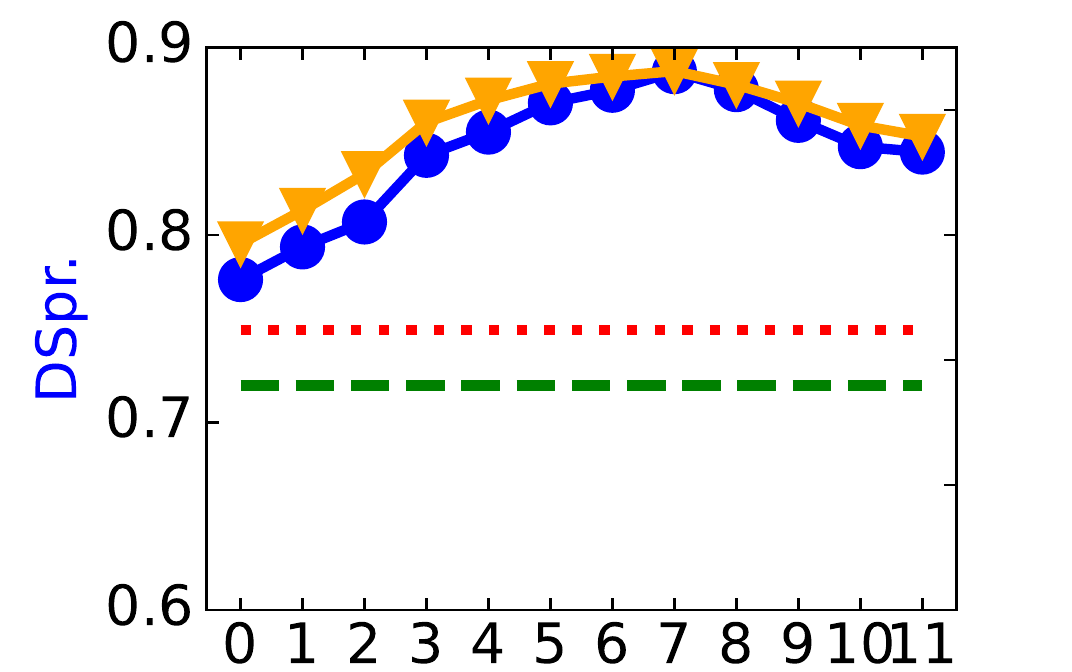}
        \caption{BERT}
        \label{fig:layer_flickr_dist_bert_val}
    \end{subfigure}
    \begin{subfigure}[t]{0.32\linewidth}
        \centering
        \includegraphics[scale=0.47]{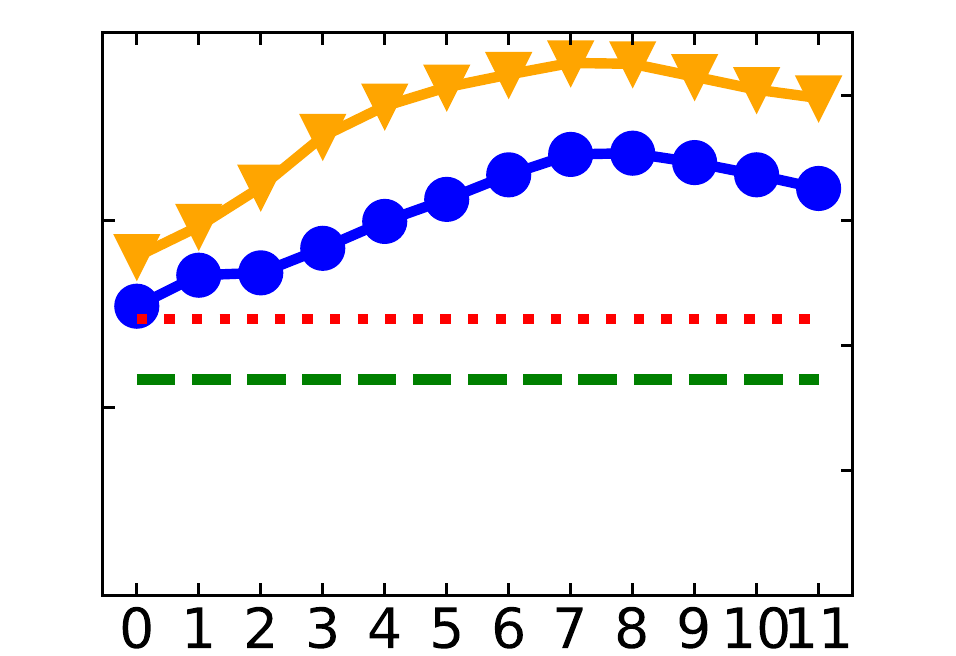}
        \caption{UNITER}
        \label{fig:layer_flickr_dist_unit_val}
    \end{subfigure}
    \begin{subfigure}[t]{0.32\linewidth}
        \centering
        \includegraphics[scale=0.47]{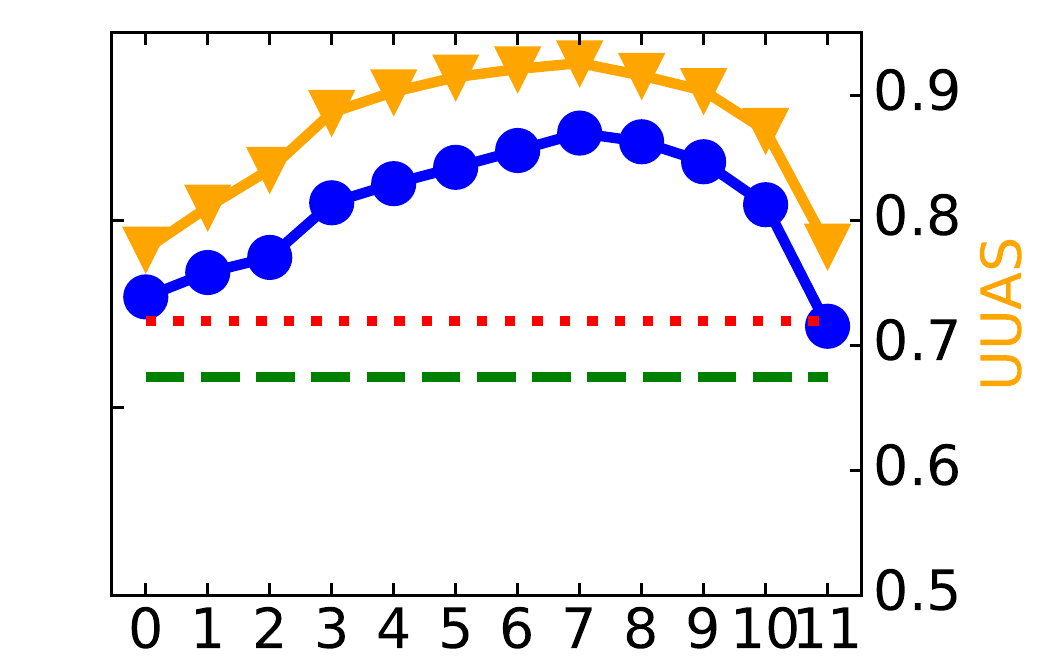}
        \caption{ViLBERT}
        \label{fig:layer_flickr_dist_vil_val}
    \end{subfigure}
    \caption{Comparison for the distance probe on the Flickr30k validation set, with textual embeddings.}
    \label{fig:layer_flickr_dist_val}
\end{figure*}
\begin{figure*}[htb]
    \centering
    \begin{subfigure}[t]{0.9\linewidth}
        \centering
        \includegraphics[scale=0.4]{figures/results/flickr30k/BERT/select_layer_ParseDepthTask_rank128.legend.pdf}
    \end{subfigure}
    \\
    \begin{subfigure}[t]{0.32\linewidth}
        \centering
        \includegraphics[scale=0.47]{figures/results/flickr30k/BERT/select_layer_ParseDepthTask_rank128.pdf}
        \caption{BERT}
        \label{fig:layer_flickr_dep_bert_val_just_text}
    \end{subfigure}
    \begin{subfigure}[t]{0.32\linewidth}
        \centering
        \includegraphics[scale=0.47]{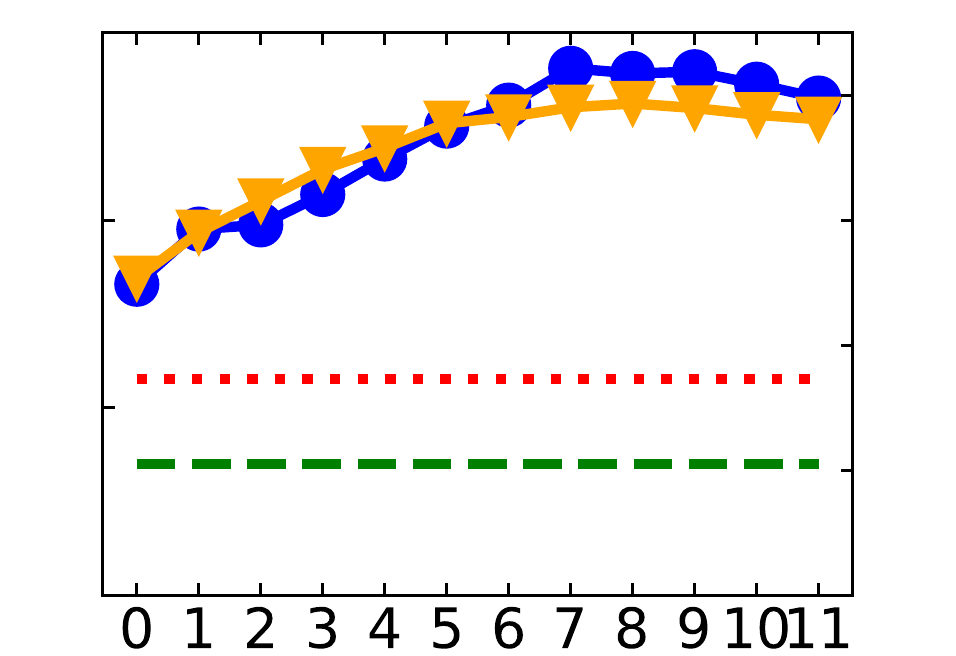}
        \caption{UNITER - only text}
        \label{fig:layer_flickr_dep_unit_val_just_text}
    \end{subfigure}
    \begin{subfigure}[t]{0.32\linewidth}
        \centering
        \includegraphics[scale=0.47]{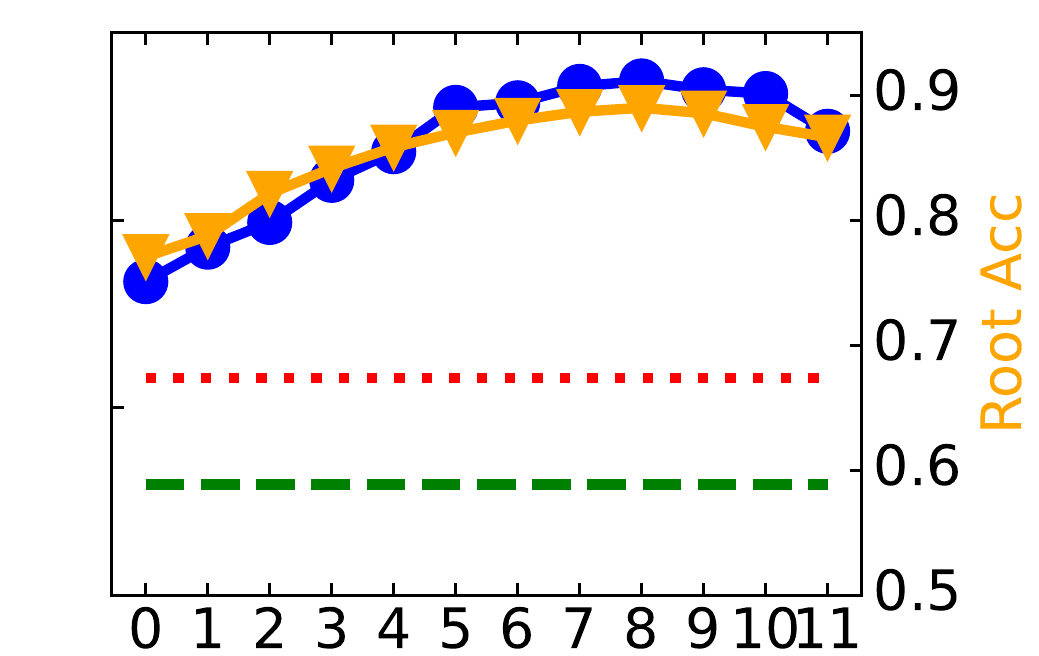}
        \caption{ViLBERT - only text}
        \label{fig:layer_flickr_dep_vil_test_just_text}
    \end{subfigure}
    \caption{Ablation comparison for the depth probe on the Flickr30k validation set while just providing textual embeddings to the multimodal-BERTs.}
    \label{fig:layer_flickr_dep_val_just_text}
\end{figure*}
\begin{figure*}[htb]
    \centering
    \begin{subfigure}[t]{0.9\linewidth}
        \centering
        \includegraphics[scale=0.4]{figures/results/flickr30k/BERT/select_layer_ParseDistanceTask_rank128.legend.pdf}
    \end{subfigure}
    \\
    \begin{subfigure}[t]{0.32\linewidth}
        \centering
        \includegraphics[scale=0.47]{figures/results/flickr30k/BERT/select_layer_ParseDistanceTask_rank128.pdf}
        \caption{BERT}
        \label{fig:layer_flickr_dist_bert_val_just_text}
    \end{subfigure}
    \begin{subfigure}[t]{0.32\linewidth}
        \centering
        \includegraphics[scale=0.47]{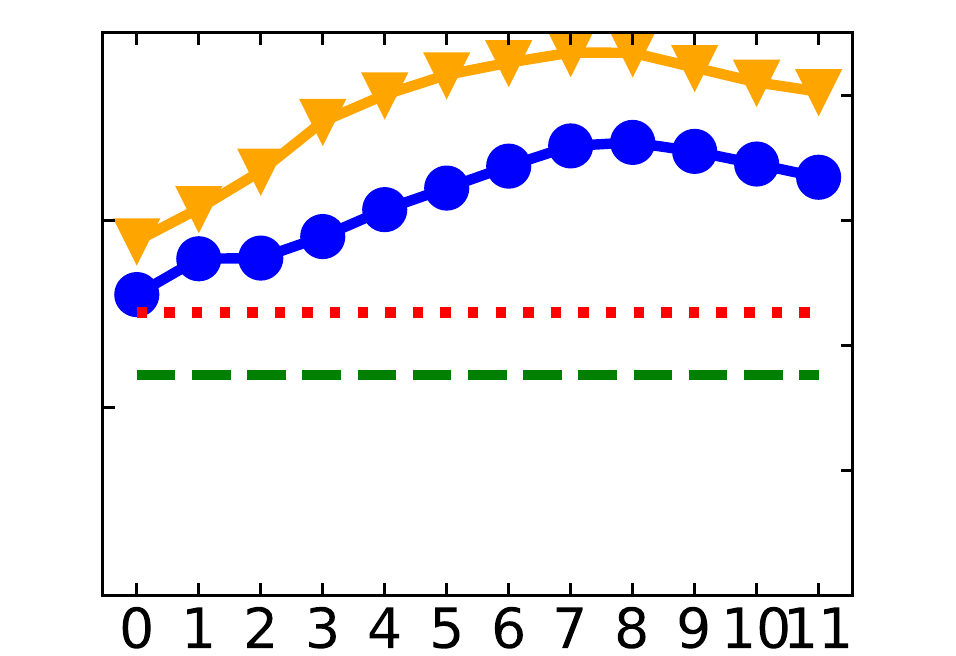}
        \caption{UNITER - only text}
        \label{fig:layer_flickr_dist_unit_val_just_text}
    \end{subfigure}
    \begin{subfigure}[t]{0.32\linewidth}
        \centering
        \includegraphics[scale=0.47]{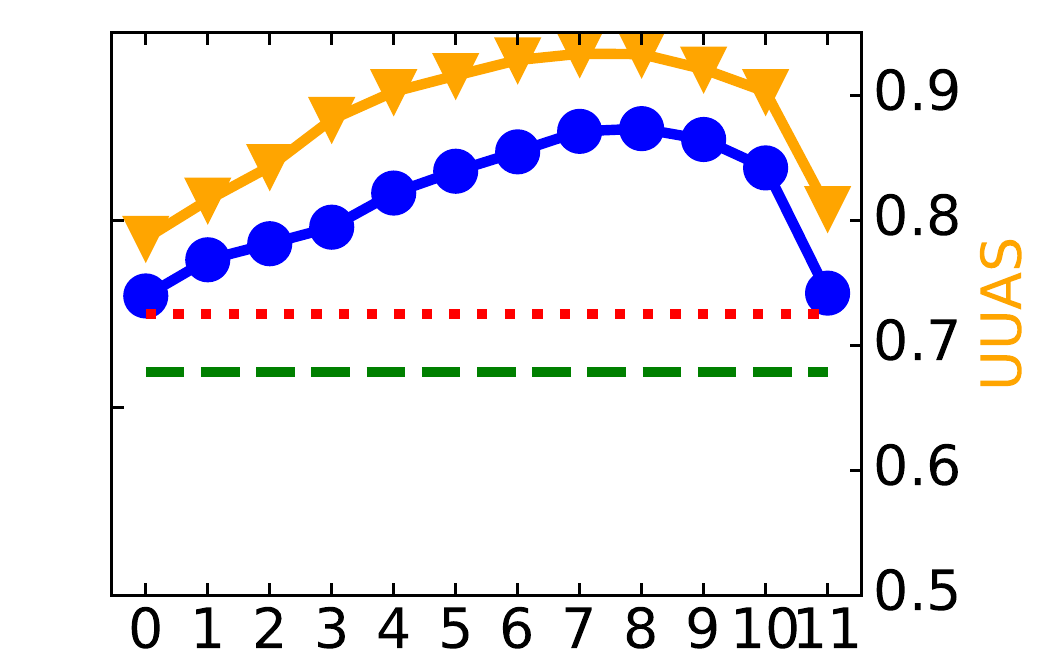}
        \caption{ViLBERT - only text}
        \label{fig:layer_flickr_dist_vil_val_just_text}
    \end{subfigure}
    \caption{Ablation comparison for the distance probe on the Flickr30k validation set while just providing textual embeddings to the multimodal-BERTs.}
    \label{fig:layer_flickr_dist_val_just_text}
\end{figure*}
\begin{figure*}[htb]
\begin{minipage}{.49\textwidth}
\flushleft
    \begin{subfigure}[t]{\linewidth}
        \centering
        \includegraphics[scale=0.4]{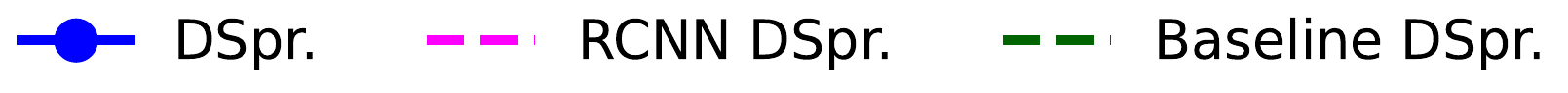}
    \end{subfigure}\\
    \begin{subfigure}[t]{0.48\linewidth}
        \flushleft
        \includegraphics[trim={0 0 0 -4},clip,scale=0.4]{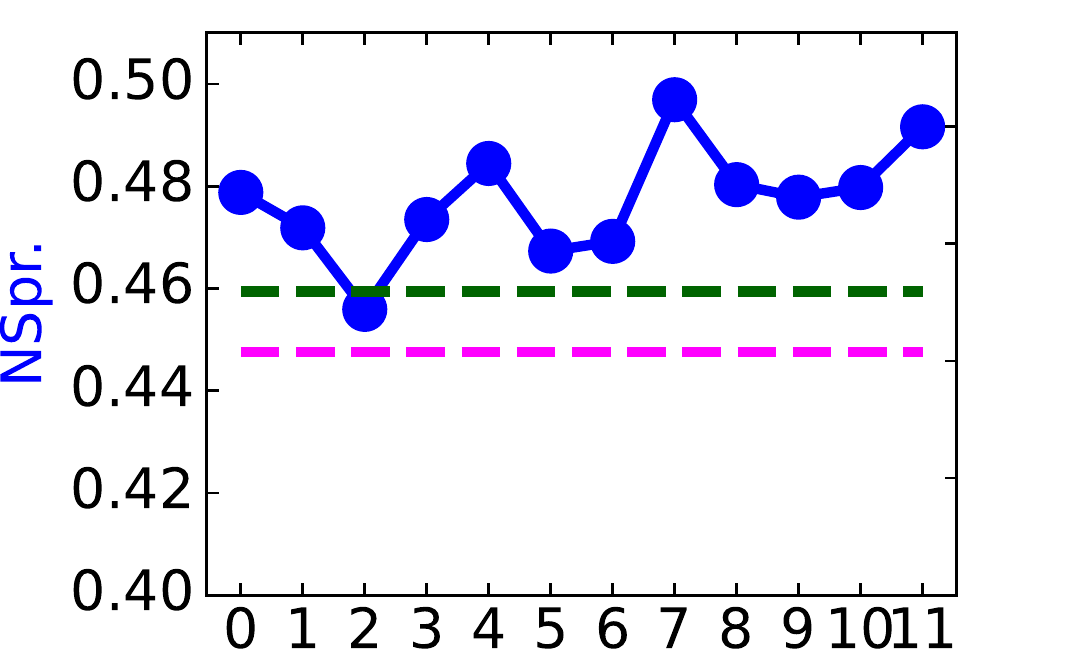}
        \caption{UNITER}
        \label{fig:layer_flickr_visdep_unit_val}
    \end{subfigure}
    \hspace{1pt}
    \begin{subfigure}[t]{0.48\linewidth}
        \flushright
        \includegraphics[trim={5 0 3 -4},clip,scale=0.4]{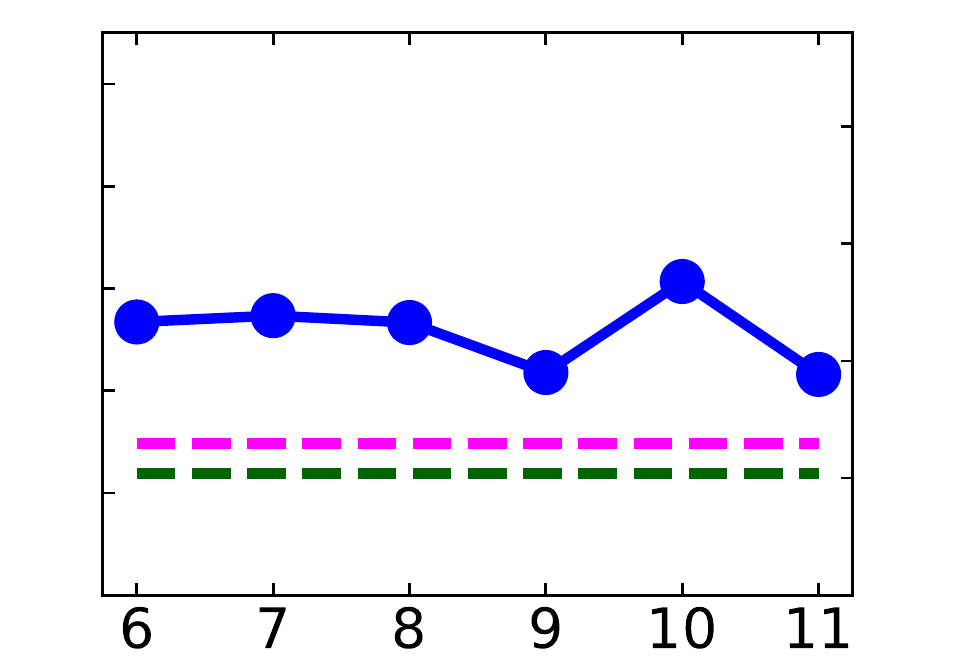}
        \caption{ViLBERT}
        \label{fig:layer_flickr_visdep_vil_val}
    \end{subfigure}
    \caption{Comparison for the depth probe on the Flickr30k validation set, with visual embeddings. Note that the scale is different in this Figure.}
    \label{fig:layer_flickr_visdep_val}
\end{minipage}
\begin{minipage}{.49\textwidth}
    \flushright
    \begin{subfigure}[t]{\linewidth}
        \centering
        \includegraphics[trim={0 10 0 3},clip,scale=0.4]{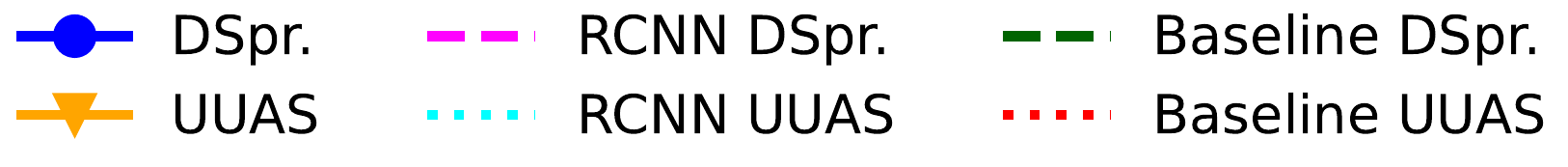}
    \end{subfigure}
    \\
    \begin{subfigure}[t]{0.47\linewidth}
        \flushleft
        \includegraphics[trim={1 0 30 0},clip,scale=0.4]{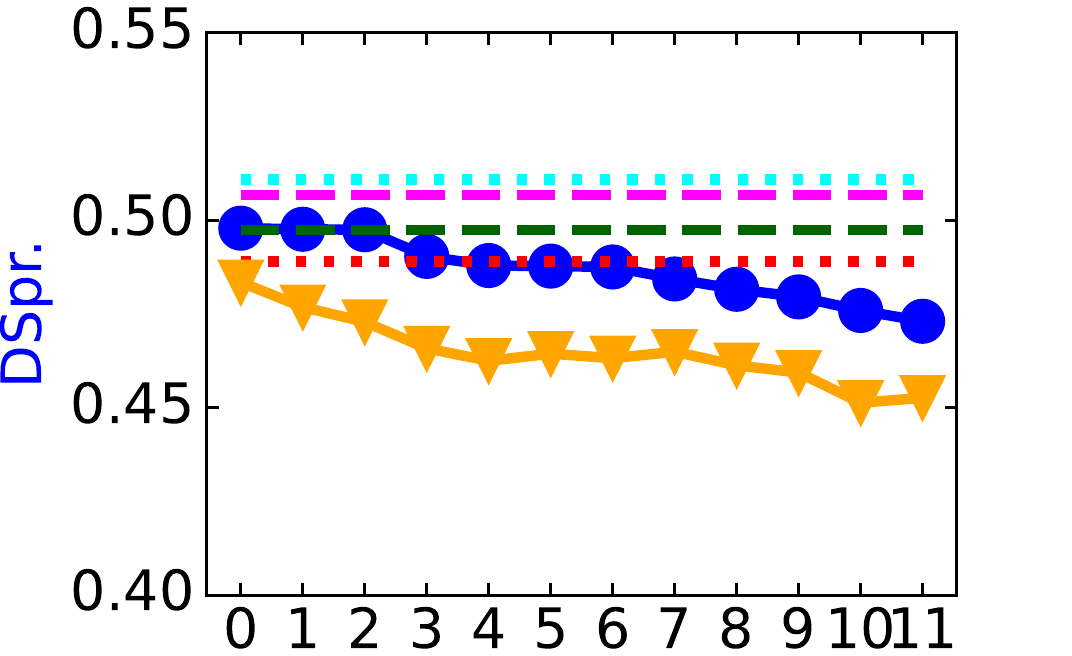}
        \caption{UNITER}
        \label{fig:layer_flickr_visdist_unit_val}
    \end{subfigure}
    ~
    \begin{subfigure}[t]{0.47\linewidth}
        \flushright
        \includegraphics[trim={30 0 0 0},clip,scale=0.4]{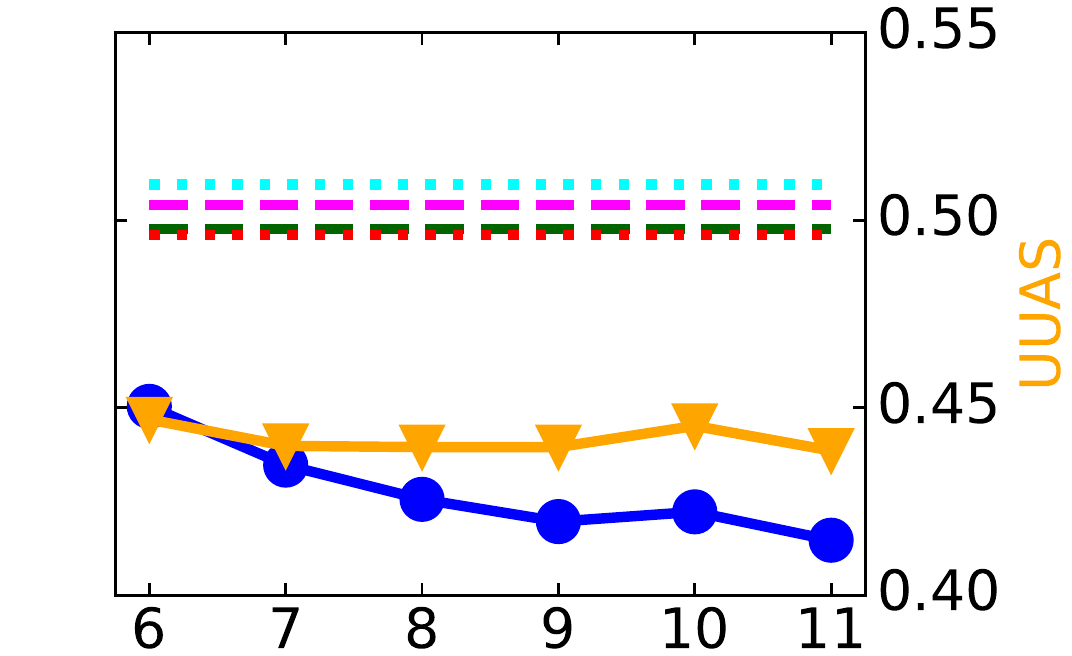}
        \caption{ViLBERT}
        \label{fig:layer_flickr_visdist_vil_val}
    \end{subfigure}
    \caption{Comparison for the distance probe on the Flickr30k validation set, with visual embeddings. Note that the scale is different in this Figure.}
    \label{fig:layer_flickr_visdist_val}
\end{minipage}
\end{figure*}
\end{document}

%% file: math_commands.tex
\def\eqref#1{equation~\ref{#1}}

\def\1{\bm{1}}

\def\vd{{\bm{d}}}

\def\vh{{\bm{h}}}

\def\mB{{\bm{B}}}

\def\mD{{\bm{D}}}

\DeclareMathAlphabet{\mathsfit}{\encodingdefault}{\sfdefault}{m}{sl}
\SetMathAlphabet{\mathsfit}{bold}{\encodingdefault}{\sfdefault}{bx}{n}

\def\sN{{\mathbb{N}}}

\def\sR{{\mathbb{R}}}

